\documentclass[11pt,a4paper]{article}

\usepackage{fontspec}
\usepackage{polyglossia}

\usepackage{tabularx}
\usepackage{makecell}
\usepackage{booktabs}

\usepackage{graphicx}
\usepackage{subcaption}

\usepackage{hyperref}

\setmainlanguage{english}
\setotherlanguage{assamese}

\usepackage{tcolorbox}
\usepackage{enumitem}

\newtcolorbox{examplebox}{
	colback=gray!5,
	colframe=gray!60,
	boxrule=0.5pt,
	arc=2mm,
	left=2mm,
	right=2mm,
	top=1mm,
	bottom=1mm
}

\newfontfamily\assamesefont[
  Script=Bengali,
  Path=./
]{NotoSerifBengali.ttf}

\usepackage{geometry}

\usepackage{natbib}
\usepackage{hyperref}

\hypersetup{
	colorlinks=true,
	linkcolor=blue!70!black,
	citecolor=blue!70!black,
	urlcolor=blue!70!black
}

\usepackage{xcolor}
\usepackage{enumitem}

\title{\textbf{AssameseLM-E2B: An Open-Weight Assamese Language Model for Efficient Consumer-Grade Deployment}}

\author{
	Pragyan Phukan\thanks{Email: phukan.pragyan@gmail.com}\\
	\small Department of Physics, Moran College, Charaideo, Assam, India, 785670
}

\date{}

\begin{document}

\maketitle

\begin{abstract}
	While Large Language Models (LLMs) excel in high-resource languages, their performance often degrades in low-resource languages due to tokenization inefficiencies, generative instability, and linguistic interference. This paper presents \textbf{AssameseLM-E2B} and its instruction-tuned variant, \textbf{AssameseLM-E2B-it}, a 2-billion-parameter model adapted specifically for low resource Assamese language. Built upon the Gemma-4 architecture, the model incorporates an expanded byte-level BPE vocabulary to reduce tokenization overhead by nearly 500\% compared with multilingual 2B--3B baselines. We employ a parameter-efficient Continual Pre-Training (CPT) pipeline with Hybrid LoRA, followed by Supervised Fine-Tuning (SFT) on a balanced, multi-domain dataset to enhance instruction-following and Retrieval-Augmented Generation (RAG) capabilities.
	
	Evaluation demonstrates a Bits-per-Byte (BPB) score of 0.7488 on FLORES-200, alongside an overall RAG score of 96.7\% and 100\% refusal accuracy on adversarial unanswerable prompts. Comparative analysis reveals significant limitations in general-purpose baselines, including receptive-productive decoupling, generative degradation on Romanized Assamese, and unintended Bengali script assimilation. These results demonstrate the potential of parameter-efficient adaptation for developing capable, locally deployable language models for low-resource languages under constrained computational resources.
\end{abstract}
%\tableofcontents

\section{Introduction}

Language modeling has undergone a paradigm shift in recent years, driven by the emergence of Large Language Models (LLMs), including OpenAI's GPT series \citep{openai2022chatgpt, openai2023gpt4}, Google's Gemini \citep{gemini2023}, and Anthropic's Claude \citep{anthropic2024claude}, alongside open-weight architectures such as LLaMA \citep{touvron2023llama}, Mistral \citep{jiang2023mistral}, Qwen \citep{qwen2024qwen25} and Gemma \citep{mesnard2024gemma}. Built upon the scalable self-attention mechanism of the Transformer architecture \citep{vaswani2017attention}, these foundational models have demonstrated remarkable capabilities in zero-shot reasoning, machine translation, and contextually coherent text generation.

Proprietary, closed-source models, such as ChatGPT, Claude, and Gemini, are typically accessed through web-based interfaces or Application Programming Interfaces (APIs). Closed-source models restrict access to their underlying source code and model weights, preventing users from independently deploying or hosting them on their own local infrastructure. These models predominantly follow an Operational Expenditure (OPEX) model, with usage-based fees generally charged on a per-token basis. Although this approach offers convenience by eliminating the need for local infrastructure and model deployment, it can incur prohibitive costs at scale. More critically, the proprietary nature of these elite systems limits mechanistic interpretability and heavily restricts open scientific research.

Conversely, open-weight architectures can be deployed on local infrastructure \citep{touvron2023llama2}, offering researchers greater control through Capital Expenditure (CAPEX). Although they require upfront investment in computational resources, they enable unrestricted inference, domain-specific adaptation, and reproducibility in local environments. Yet, despite the democratizing promise of open models, an overwhelming English-centric bias persists in modern foundational architectures. For instance, an empirical breakdown of expansive pre-training corpora reveals that high-resource European languages account for nearly 99.7\% of total tokens, whereas Indic languages and other underrepresented languages collectively constitute a negligible fraction (frequently under 0.21\%).

Recent initiatives have made significant progress in addressing the linguistic divide in the Indian language ecosystem. Frameworks such as IndicTrans2 \citep{gala2023indictrans2} provide high-quality neural machine translation across  22  Indian languages, while Airavata \citep{gala2024airavata} adapts open foundational architectures for Hindi instruction following. Similarly, Tamil-LLaMA \citep{balachandran2023tamilllama} demonstrates the effectiveness of vocabulary expansion for enhancing language-specific capabilities. More recent efforts include BharatGen's Param1-7B, a multilingual Mixture-of-Experts (MoE) model incorporating Assamese and 13 other Indian languages alongside English and Hindi \citep{bharatgen2025param1}, and Sarvam AI's Indic-focused language models, including Sarvam-M, which demonstrate the potential of specialized post-training for improving multilingual reasoning and generation across Indian languages \citep{sarvamai2024sarvam}. Despite these advances, substantial disparities persist in the representation of India's linguistic diversity. In particular, Assamese and several other languages of Northeast India remain comparatively underrepresented in large-scale pretraining corpora, language-specific evaluation benchmarks, and specialized generative AI research.

Applying off-the-shelf multilingual LLMs to Assamese introduces two fundamental computational and linguistic failures:

\paragraph{The Assamese Tokenization Bottleneck and the ``Token Tax":} State-of-the-art tokenizers are trained on corpora overwhelmingly dominated by Latin scripts \citep{ahia2023languages}. Because native Assamese subwords are virtually absent from base vocabularies, tokenizers fall back on raw byte-level encodings, frequently decomposing a single Assamese character into 3 to 4 individual UTF-8 byte tokens. This produces an excessively high subword fertility rate, bloating input sequence lengths by 300\% to 500\% \citep{rust2021tokenizer}. This structural inefficiency slows inference throughput by a factor of three, rapidly exhausts the model's finite context window, and levies an exorbitant Token Tax'' on downstream users of commercial APIs who must pay per generated subword token \citep{ahia2023languages}.

\paragraph{Cross-Lingual Bengali Contamination:} Assamese utilizes the Eastern Indic (Bengali-Assamese) script, sharing significant orthographic overlap with Bengali \citep{bhattacharyya2021challenges}. However, because Bengali has a vastly larger digital footprint on the open web, generic multilingual models trained on uncurated web dumps suffer from severe cross-lingual contamination. When prompted in Assamese, these models routinely hallucinate Bengali vocabulary, adopt Bengali morphological conjugations, and fail to recognize distinct Assamese graphemes, specifically the characters \textbf{{\assamesefont ৰ}} (\textit{ra}) and \textbf{{\assamesefont ৱ}} (\textit{wa}) . Standard tokenizers conflate the two distributions, effectively forcing language models to treat Assamese as low-quality or corrupt Bengali.

To overcome these structural limitations, we introduce \textbf{AssameseLM-E2B}, a 2-billion parameter Language Model (LM) natively adapted for the Assamese language. Rather than pre-training a massive 70-billion parameter system, we deliberately ground our work on the lightweight, high-throughput Gemma-4-E2B architecture \citep{mesnard2024gemma}. At \(\sim\)2B parameters, AssameseLM-E2B is fully edge-ready: it circumvents the prohibitive multi-GPU infrastructure required by larger models, democratizing local inference, fine-tuning, and deployment on standard consumer-grade hardware.

Drawing inspiration from vocabulary expansion strategies demonstrated in Chinese-LLaMA \citep{cui2023chinese} and Tamil-LLaMA \citep{balachandran2023tamilllama}, we augment the base Gemma tokenizer with an additional, carefully curated Assamese vocabulary. To train the expanded embeddings and adapt the attention layers efficiently without catastrophic forgetting or insurmountable compute budgets, we employ the Low-Rank Adaptation (LoRA) methodology \citep{hu2021lora}. We perform Continual Pre-Training (CPT) on an Assamese corpus specifically curated and filtered to eliminate Bengali linguistic contamination, resulting in the open-weight model \textbf{AssameseLM-E2B}\footnote{\url{https://huggingface.co/assamailab/AssameseLM-E2B}}. Subsequently, we conduct Supervised Fine-Tuning (SFT) using instruction-following datasets to develop its instruction-tuned counterpart, \textbf{AssameseLM-E2B-it} \footnote{\url{https://huggingface.co/assamailab/AssameseLM-E2B-it}}, which is also released as an open-weight model. The fine-tuned model is designed to perform effectively in Retrieval-Augmented Generation (RAG) environments, making it a practical and cost-effective choice for academic institutions and small-scale industries. Its lightweight architecture allows deployment on local infrastructure or general-purpose cloud servers, reducing dependence on proprietary, API-based services and their associated per-token usage costs.

The principal contributions of this paper are summarized as follows:
\begin{itemize}[leftmargin=*]
    \item \textbf{Assamese Vocabulary Expansion:} We train a custom SentencePiece tokenizer \citep{kudo2018sentencepiece} on native Assamese text and merge it with the base Gemma tokenizer. This expansion significantly reduces Assamese token fertility, allowing the model to handle three times more information in its context window and process sequences significantly faster while utilizing less memory.
    \item \textbf{Contamination-Filtered Pre-Training Corpus:} We curate an Assamese pre-training dataset combining clean web-scraped corpora with historical, literary, and educational texts digitized through an advanced Optical Character Recognition (OCR) pipeline, strictly isolated from Bengali lexical contamination.
  \item \textbf{Parameter-Efficient Continual Pre-Training:} We perform Continual Pre-Training (CPT) on the Gemma-4-E2B base model using Low-Rank Adaptation (LoRA) \citep{hu2021lora}. This process adapts the newly expanded embedding layer and attention components to capture native Assamese syntax, semantics, and morphology efficiently. The resulting foundational model is released under the name AssameseLM-E2B.
\item \textbf{Instruction Fine-Tuning for Conversational Alignment:} We develop a targeted Assamese instruction-following dataset by adapting translated subsets of Alpaca \citep{taori2023alpaca} and OpenOrca \citep{lian2023openorca}, supplemented with culturally relevant, localized prompts. Using this dataset, we perform Supervised Fine-Tuning (SFT) on AssameseLM-E2B to enhance its instruction-following capabilities and conversational alignment. The resulting instruction-tuned model is released under the name AssameseLM-E2B-it.
    \item \textbf{Open-Source Democratization:} To foster transparent academic inquiry and empower developers, we release both the foundational base model weights (AssameseLM-E2B), the instruction-tuned chat model (AssameseLM-E2B-it) under an open license.
\end{itemize}

The remainder of this paper is structured as follows. Section 2 provides the necessary background on the foundational Gemma model architecture. Section 3 introduces the AssameseLM-E2B base model, detailing our methodology for expanding the Assamese vocabulary, the curation of high-fidelity pre-training datasets, the parameter-efficient continual pre-training phase and quantitative evaluation on FLORES-200 Benchmark. Section 4 presents AssameseLM-E2B-it, our instruction-tuned conversational agent. This section outlines the supervised fine-tuning dataset curation, the training methodology, and the optimization dynamics, culminating in a comprehensive empirical evaluation of both Retrieval-Augmented Generation (RAG) and multi-dimensional direct chat instruction capabilities. Section 5 provides a summary of our contributions and the future outlook for low-resource language modeling. Finally, Section 6 supplies the appendix resources, including links to our open-source weights and datasets and Section 7 showcases qualitative inference examples.

\section{Background on Gemma Models}

The Gemma family of open-weights language models, developed by Google, shares its foundational architectural DNA and pre-training methodologies with the flagship Gemini models \citep{mesnard2024gemma}. Designed to offer state-of-the-art performance within highly constrained computational budgets, Gemma models are particularly well-suited for deployment on consumer-grade hardware and edge devices.

For this research, we selected the \textbf{Gemma-4-E2B} (2-billion parameter) variant as our foundational backbone. Architecturally, Gemma utilizes a decoder-only transformer network equipped with Rotary Position Embeddings (RoPE), GeGLU activation functions, and RMSNorm for layer normalization. The 2B variant is specifically optimized for efficiency, making it an ideal candidate for Continual Pre-Training (CPT) and Parameter-Efficient Fine-Tuning (PEFT) pipelines via Low-Rank Adaptation (LoRA) \citep{hu2021lora}. Furthermore, its native compatibility with Scaled Dot-Product Attention (SDPA) allows for significant memory optimization without the hardware-specific constraints often associated with FlashAttention implementations. However, in its base multilingual form, Gemma's tokenizer remains heavily biased toward high-resource languages, leading to extreme subword fragmentation when processing the Eastern Nagari script of the Assamese language. Addressing this structural limitation was the primary motivation for our vocabulary expansion and domain-adaptation methodology.

\section{AssameseLM-E2B}
\subsection{Expansion of Assamese Vocabulary}

\begin{table}[h!]
	\centering
	\caption{Tokenizer comparison between original Gemma-4-E2B and AssameseLM-E2B}
	\label{table:1}
	\begin{tabularx}{\textwidth}{c c X c}
		\hline
		& \textbf{Length} & \textbf{Content} &\textbf{\makecell{Tokens per\\Word Ratio}} \\
		\hline
		Assamese Text & \makecell{16\\(Word Count)} &
		\textbf{\assamesefont শিক্ষা হৈছে মানৱ জীৱনৰ আটাইতকৈ গুৰুত্বপূৰ্ণ সম্পদ, যিয়ে অজ্ঞানতাৰ আন্ধাৰ দূৰ কৰি জ্ঞানৰ পোহৰ বিলাই দিয়ে।}
		&  \\
		\hline
		Gemma-4-E2B & \makecell{43\\(Tokens Count)}  & \textbf{\assamesefont 'শিক্ষা', ' হৈ', 'ছে', ' মান', 'ৱ', ' জী', 'ৱ', 'ন', 'ৰ', ' আ', 'টাই', 'ত', 'ক', 'ৈ', ' গু', 'ৰ', 'ু', 'ত্ব', 'পূ', 'ৰ', '্ণ', ' সম্পদ', ',', ' য', 'িয়ে', ' অজ্ঞান', 'তা', 'ৰ', ' আ', 'ন্ধ', 'াৰ', ' দূ', 'ৰ', ' কৰি', ' জ্ঞান', 'ৰ', ' প', 'োহ', 'ৰ', ' বি', 'লাই', ' দিয়ে', '।' }& 2.7 \\
		\hline
		AssameseLM-E2B & \makecell{24\\(Tokens Count)}   & \textbf{\assamesefont 'শিক্ষা', ' হৈছে', ' মানৱ', ' জীৱনৰ', ' আটাইতকৈ', ' গুৰুত্বপূৰ্ণ', ' ', 'সম্পদ', ',', ' যিয়ে',  অজ', '্', 'ঞা', 'নতা', 'ৰ', ' আন্ধাৰ', ' দূৰ', ' ', 'কৰ', 'ি', ' জ্ঞানৰ', ' পোহৰ', ' বিলাই', ' দিয়ে।'} & 1.5 \\
		\hline
	\end{tabularx}
\end{table}

As a morphologically rich language, Assamese presents unique tokenization challenges for existing foundational models. The default vocabulary of foundational models like Gemma severely underrepresents the Eastern Indic script, leading to excessive subword fragmentation during encoding. To resolve this structural inefficiency, we expanded the base tokenizer with natively derived Assamese subwords, drawing inspiration from vocabulary extension methodologies utilized in Chinese-LLaMA \citep{cui2023chinese} and Tamil-LLaMA \citep{balachandran2023tamilllama}.

To ensure architectural compatibility with the base Gemma model, we utilized the SentencePieceBPETokenizer algorithm, relying on the Byte-Pair Encoding (BPE) framework established by \citet{kudo2018sentencepiece}. The custom tokenizer was trained entirely on our curated monolingual Assamese corpus, which included both aggregated external datasets and our proprietary Continual Pre-Training (CPT) texts. We initialized the training pipeline with a target vocabulary size of 20,000 tokens. To systematically exclude typographical anomalies and statistical noise from the learned vocabulary, a minimum token frequency threshold of 2 was enforced during generation.

A critical challenge in applying unconstrained BPE algorithms to complex Indic scripts is the generation of orthographically invalid subwords. In the Assamese script, dependent vowel signs (matras), modifier markers such as Chandrabindu (\textbf{{\assamesefont ঁ}}),  Anusvara (\textbf{{\assamesefont ং}}) and Visarga (\textbf{{\assamesefont ঃ}})), and the Virama / Hasant (\textbf{{\assamesefont ্}}) are orthographically dependent; they cannot linguistically exist at the beginning of a standalone word or valid subword token. If injected into the model's embedding layer, these ``broken'' graphemes force the language model to learn corrupted morphological representations.

To prevent this, we implemented a rigorous algorithmic filtering heuristic post-BPE training:

\begin{description}[leftmargin=!]
	\item[\textbf{Unicode Isolation:}] We applied a regular expression filter to retain only tokens containing valid characters within the Eastern Indic Unicode block (\texttt{\textbackslash u0980}--\texttt{\textbackslash u09FF}), stripping out pure English or noise tokens generated during mixed-corpus processing.
	\item[\textbf{Invalid Grapheme Exclusion:}] We implemented a strict rule-based exclusion matrix to drop any token that improperly commenced with an isolated modifier. Specifically, the script discarded any generated subword beginning with dependent vowel signs (e.g.,  \textbf{{\assamesefont া, ি, ী, ু}}), Chandrabindu (\textbf{{\assamesefont ঁ}}),  Anusvara (\textbf{{\assamesefont ং}}), Visarga (\textbf{{\assamesefont ঃ}}), or the Virama/Hasant (\textbf{{\assamesefont ্}}), even when prefixed by the SentencePiece meta-space character (\texttt{U+2581}).
\end{description}

This deterministic filtering process yielded a highly sanitized, linguistically accurate set of native Assamese tokens. Following established vocabulary amalgamation practices, these clean tokens were appended to the original Gemma tokenizer. Subsequently, the base model’s word embedding matrix and language model head were dynamically resized from their original shape $V \times H$ to an expanded shape $V' \times H$, where $V'$ represents the new aggregated vocabulary size. This expansion strategy preserves the pre-trained weights of the original English vocabulary while drastically reducing the token-per-word ratio and computational overhead.

\subsection{Datasets Curation for Pre-training}

The efficacy of Continual Pre-Training (CPT) heavily depends on the linguistic diversity, grammatical integrity, and volume of the underlying corpus \citep{bhattacharyya2021challenges}. A fundamental challenge in Assamese NLP is that standard, open-source web crawls are frequently riddled with code-switching, grammatical errors, and cross-lingual contamination \citep{bhattacharyya2021challenges}. To establish a robust linguistic foundation for AssameseLM-E2B, we curated a hybrid dataset that balances the massive token volume of open-source web dumps with the high grammatical quality of targeted, proprietary datasets.

The dataset curation process was structured across the following three streams:

\begin{enumerate}[label=\textbf{\roman*.}, leftmargin=*]
	\item \textbf{High-Fidelity OCR Literary Corpus:} Although large-scale web crawls provide the substantial token volume required for language modeling, they often lack the grammatical consistency and linguistic quality of formally published texts. To address this limitation, we digitized approximately 100 books, comprising Assamese-medium textbooks, classical literature, and popular novels, using an advanced Optical Character Recognition (OCR) pipeline to enrich the corpus with grammatically reliable and linguistically diverse content. Although this proprietary subset constitutes a comparatively small fraction of the total pre-training token volume, it serves as an indispensable linguistic anchor. Unlike unstructured internet data, these digitized books inject highly formal Assamese grammar, rich historical and literary vocabulary, and complex sentence structures into the model.
	
	\item \textbf{Contemporary News Scraping:} Because formal literature and generic web dumps often suffer from temporal staleness, we engineered custom web scrapers to systematically extract articles from major Assamese daily news portals. This localized dataset bridges the gap between classic literature and internet slang. It grounds the model in modern vernacular, evolving journalistic vocabulary, and contemporary socio-political contexts specific to Assam, ensuring the language model understands current cultural discourse.
	
	\item \textbf{Large-Scale Web and Encyclopedic Corpora:} To supply the massive token volume required to stabilize Continual Pre-Training, we integrated several standard open-source corpora. This included the Assamese subsets of the Common Crawl corpus and the OSCAR dataset \citep{caswell2020language}, which provide a vast breadth of unstructured internet text. Furthermore, we incorporated the complete Assamese Wikipedia dump to inject structured, factual, and encyclopedic reasoning, alongside the Sangraha Assamese corpus \citep{ai4bharat2023sangraha}, a specialized Indic dataset curated specifically for regional language modeling.
\end{enumerate}

\subsubsection*{Data Sanitization and Deduplication:} 
Because aggregating proprietary OCR text, news scrapes, and massive web crawls inevitably introduces artifacts and redundancy, a rigorous multi-stage sanitization pipeline was enforced:

\begin{enumerate}[label=\textbf{\roman*.}, leftmargin=*]
	\item\textbf{OCR Error Correction:} Heuristic filters were applied specifically to our digitized book corpus to correct common OCR artifacts, such as misread conjunct consonants, hyphenation errors, and broken margins.
	\item\textbf{Deduplication:} We executed exact-match and MinHash-based fuzzy deduplication \citep{broder1997resemblance, lee2021deduplicating} across the merged corpora to eliminate repetitive boilerplate text, duplicate news syndications, and overlapping Wikipedia entries.
	\item\textbf{Contamination Filtering:} Finally, strict Unicode-level filtering was applied to strip out English HTML remnants and cross-lingual Bengali contamination, ensuring the final pre-training corpus was a highly dense, uncontaminated, and structurally accurate representation of the Assamese language.
\end{enumerate}

\subsection{Pre-training Phase}

To harness the newly expanded Assamese vocabulary and adapt the base model to the structural nuances of the Assamese language, we executed a Continual Pre-Training (CPT) on Gemma-4-E2B foundational model. The primary objective of this phase is Causal Language Modeling (CLM), wherein the model learns the semantic and syntactic distributions of the Assamese language by predicting the next token in a sequence based on the preceding context.

Given the computational constraints of fully retraining a 2-billion parameter model, we employed Parameter-Efficient Fine-Tuning (PEFT) utilizing the Low-Rank Adaptation (LoRA) methodology \citep{hu2021lora}. For the training process, we utilized an NVIDIA RTX A6000 GPU equipped with 48 GB of VRAM , supported by 50 GB of system RAM and 9 vCPUs. The model was trained for one epoch over the entire dataset.

\subsubsection*{Hybrid LoRA Architecture} 
To ensure compatibility between the underlying Gemma-4 architecture \citep{mesnard2024gemma} and the PEFT library, we dynamically modified the model graph by replacing the proprietary \texttt{Gemma4ClippableLinear} modules with standard \texttt{nn.Linear} layers prior to LoRA initialization. To maximize the model's adaptive capacity, we targeted all primary linear projection matrices across the attention mechanisms and feed-forward networks (\texttt{q\_proj}, \texttt{k\_proj}, \texttt{v\_proj}, \texttt{o\_proj}, \texttt{gate\_proj}, \texttt{up\_proj}, and \texttt{down\_proj}).

Crucially, because we expanded the tokenizer with native Assamese subwords, standard LoRA, which exclusively trains adapter weights, would fail to learn representations for the newly added tokens. To resolve this, we implemented a ``Hybrid LoRA'' approach by explicitly unfreezing the input embedding layer (\texttt{embed\_tokens}) and the final classification head (\texttt{lm\_head}). This ensured that the model comprehensively learned the semantic weights of the new Assamese vocabulary alongside the adapted attention mechanisms.

\subsubsection*{Training Hyperparameters and Optimization} 
The Continual Pre-Training (CPT) phase was conducted using the \texttt{SFTTrainer} class from the Transformer Reinforcement Learning (TRL) library. We loaded the base model using \texttt{bfloat16} precision and accelerated attention computations via Scaled Dot-Product Attention (SDPA).

To optimize GPU memory utilization during training, we enabled gradient checkpointing. Furthermore, because our pre-training corpus included unstructured text from OCR transcriptions and web scrapes, we enabled sequence packing. Packing concatenates multiple shorter documents into single dense blocks of the maximum sequence length, drastically improving training throughput and eliminating padding inefficiencies. The complete set of optimization constraints and training hyperparameters is detailed in Table~\ref{tab:cpt_hyperparams}.

\begin{table}[h]
	\centering
	\caption{Hyperparameters for the Continual Pre-Training (CPT) Phase}
	\label{tab:cpt_hyperparams}
	\begin{tabular}{@{}ll@{}}
		\toprule
		\textbf{Hyperparameter} & \textbf{Value} \\
		\midrule
		LoRA Rank ($r$) & 64 \\
		LoRA Alpha ($\alpha$) & 128 \\
		LoRA Target Modules &\texttt{ \makecell{q\_proj, k\_proj, v\_proj, o\_proj,\\ gate\_proj, up\_proj, down\_proj}} \\
		LoRA Dropout & 0.05 \\
		Precision & \texttt{bfloat16} \\
		Max Sequence Length & 2048 \\
	%	Sequence Packing & True \\
		Optimizer & \texttt{adamw\_torch} \\
		Learning Rate & $2 \times 10^{-4}$ \\
	%	Learning Rate Scheduler & Cosine \\
	%	Warmup Steps & 100 \\
	%	Per-Device Batch Size & 4 \\
	%	Gradient Accumulation Steps & 8 \\
		Training Epochs & 1 \\
		\bottomrule
	\end{tabular}
\end{table}

\subsubsection*{Pre-training Convergence and Training Dynamics}

\begin{figure}[htbp]
	\centering
	
	\begin{subfigure}{0.5\textwidth}
		\centering
		\includegraphics[width=\linewidth]{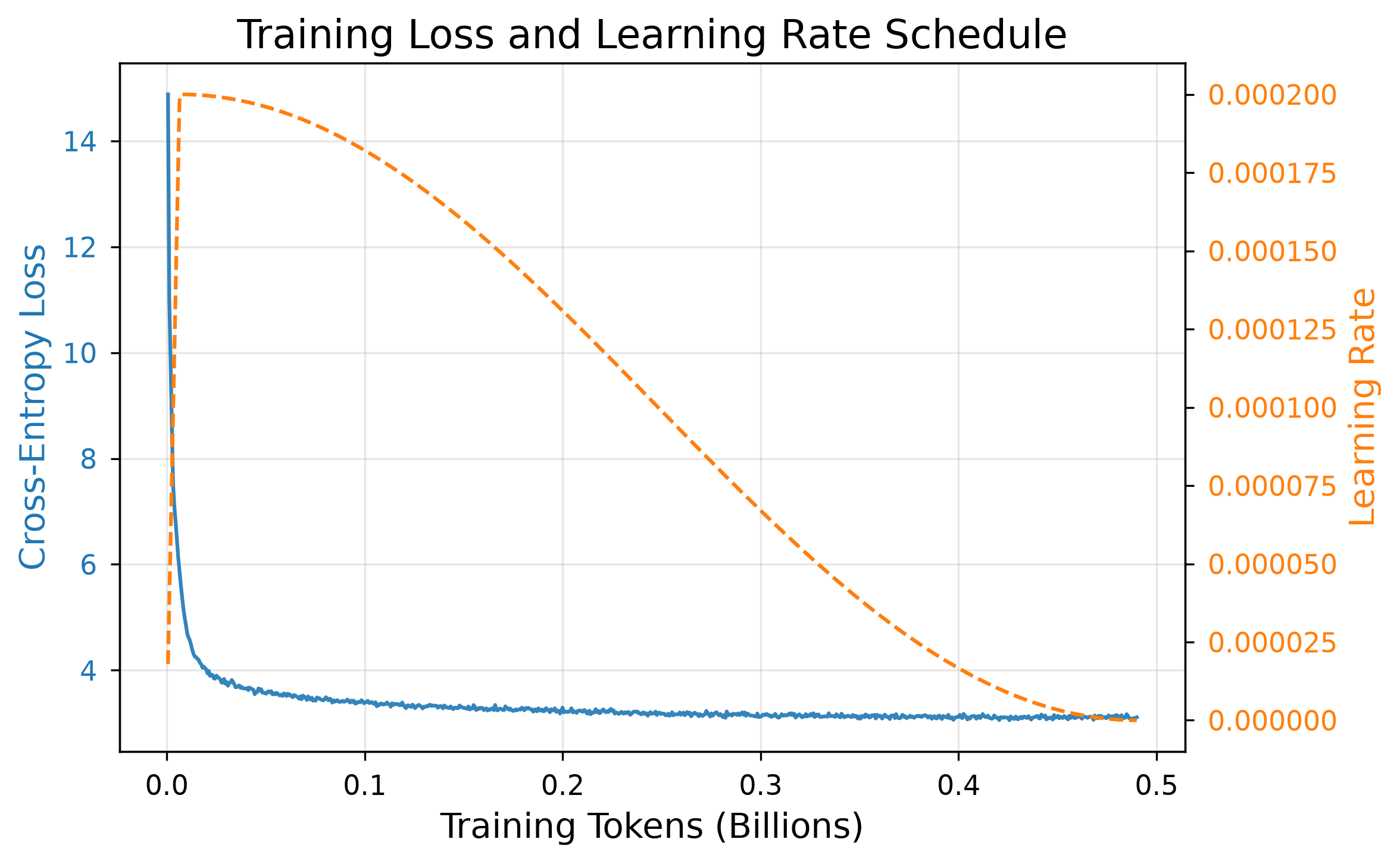}
		\caption{Training Tokens vs.TL \& LR}
		\label{fig:image1}
	\end{subfigure}
	\hfill
	\begin{subfigure}{0.45\textwidth}
		\centering
		\includegraphics[width=\linewidth]{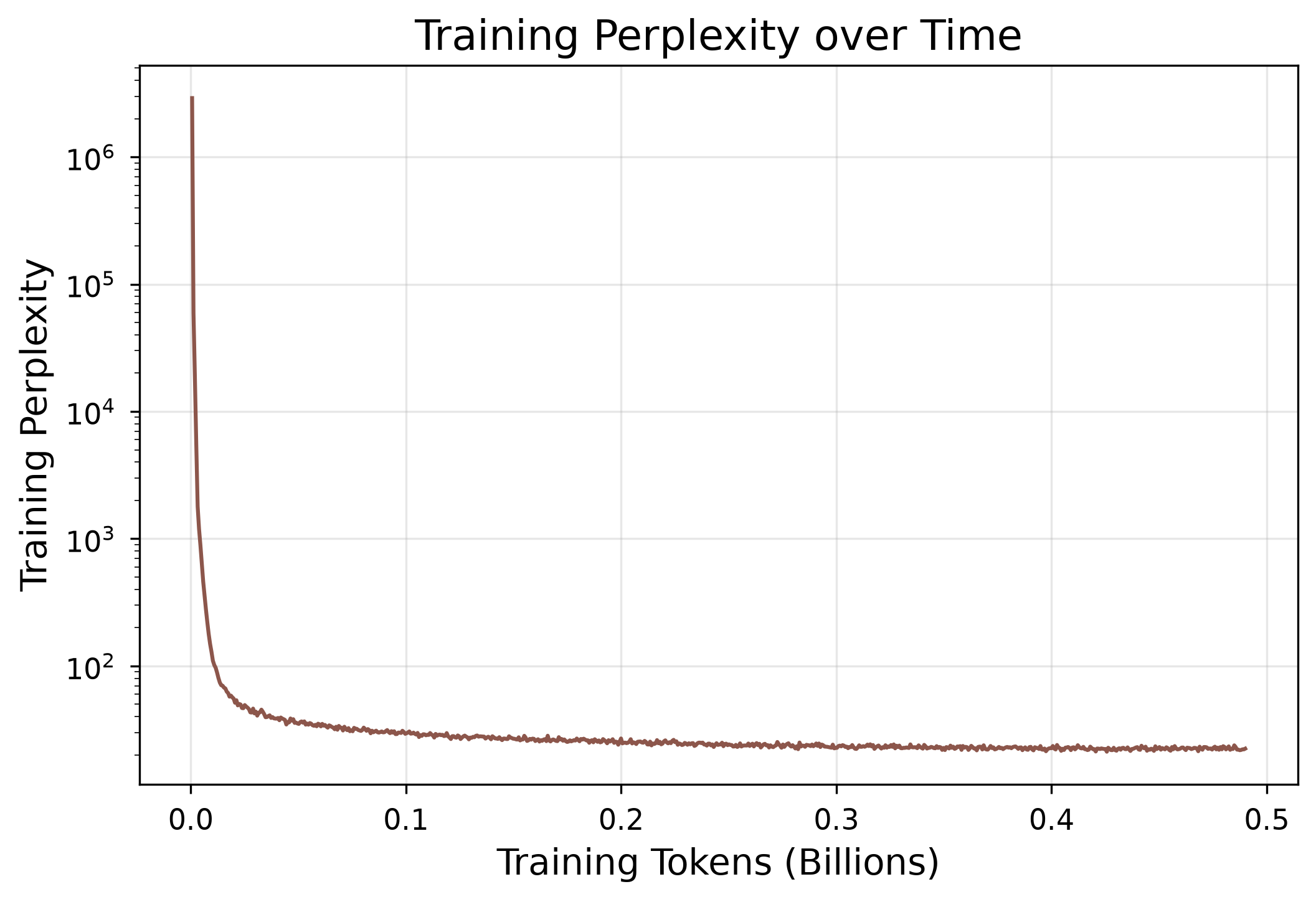}
		\caption{Training Tokens vs Training Perplexity}
		\label{fig:image2}
	\end{subfigure}\\
	\vspace{0.5cm} 
	\begin{subfigure}{0.32\textwidth}
		\centering
		\includegraphics[width=\linewidth]{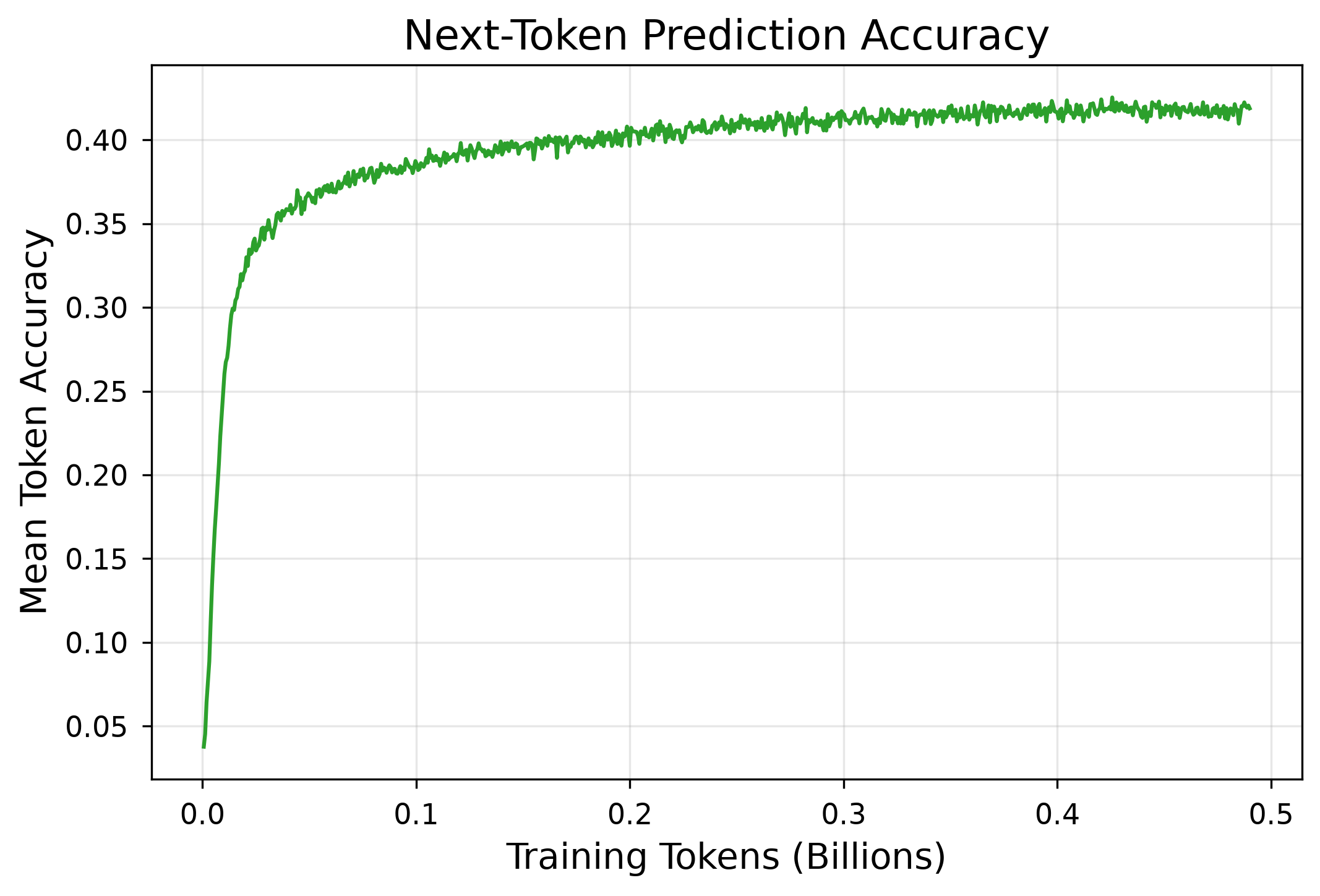}
		\caption{Training Tokens vs. Next-Token Prediction Accuracy}
		\label{fig:image3}
	\end{subfigure}
	\hfill
	\begin{subfigure}{0.32\textwidth}
		\centering
		\includegraphics[width=\linewidth]{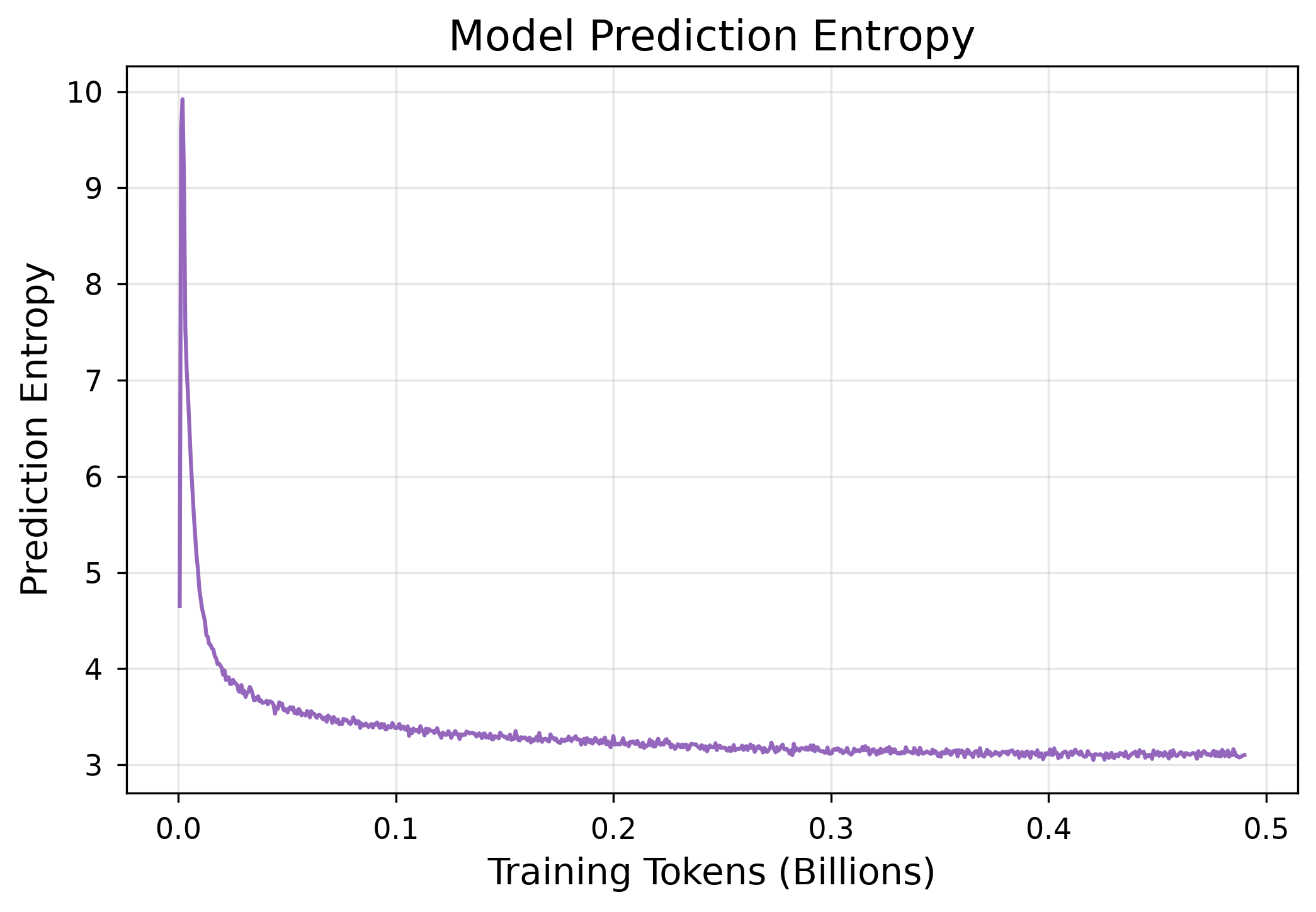}
		\caption{Training Tokens vs. Model Prediction Entropy}
		\label{fig:image4}
	\end{subfigure}
		\hfill
	\begin{subfigure}{0.32\textwidth}
		\centering
		\includegraphics[width=\linewidth]{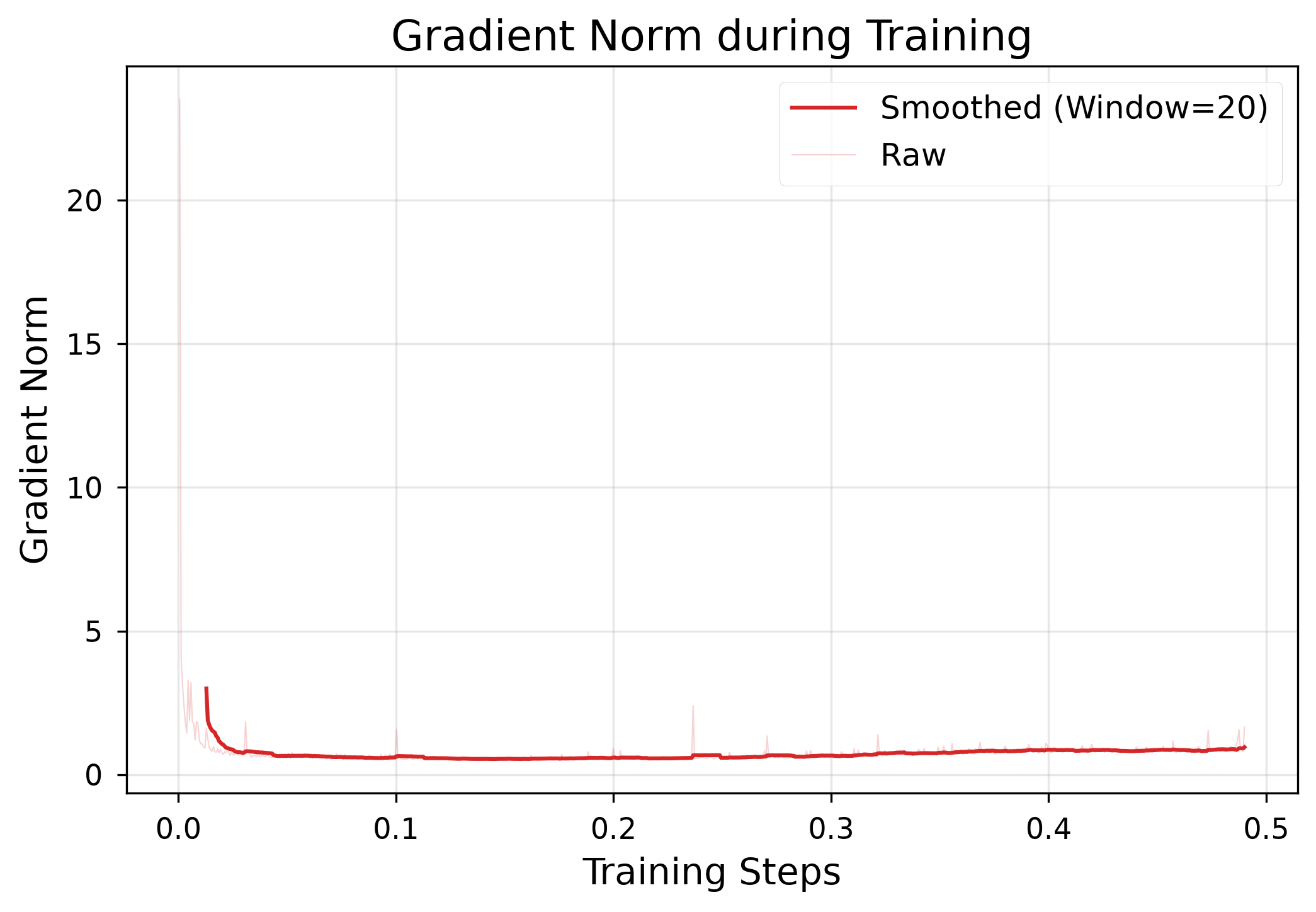}
		\caption{Training Steps vs. Gradient Norm}
		\label{fig:image5}
	\end{subfigure}
	
	\caption{CPT Training Dynamics of AssameseLM-E2B}
	\label{fig:four_images}
\end{figure}

The Continual Pre-Training (CPT) phase exhibited highly stable and rapid convergence over a corpus of 0.5 billion Assamese tokens. The cross-entropy loss initialized at an exceptionally high value of approximately 15 (Figure~\ref{fig:image1}). Concurrently, the initial training perplexity exceeded 106 on a logarithmic scale (Figure~\ref{fig:image2}). This initial divergence is mathematically expected; because the model's newly expanded Assamese vocabulary embeddings were randomly initialized, the earliest forward passes produced essentially uniform, non-semantic logits. However, guided by a cosine learning rate scheduler that peaked at a maximum learning rate of 0.000200 (Figure~\ref{fig:image1}), the model demonstrated immediate adaptation. Within the first 0.05 billion tokens, the cross-entropy loss dropped precipitously. The loss curve subsequently smoothed, gradually decaying to a final plateau near a value of 3.1 by the end of the 0.5 billion token training run. Correspondingly, training perplexity collapsed from its initial peak and stabilized firmly between 101 and 102 (Figure~\ref{fig:image2}), indicating that the model successfully mapped the new Assamese subwords into a coherent latent semantic space.\\

The progressive linguistic alignment of the model is further validated by the next-token prediction accuracy (Figure~\ref{fig:image3}) and model prediction entropy metrics (Figure~\ref{fig:image4}). As we can see from the (Figure~\ref{fig:image4}), at the onset of training, the mean token accuracy was virtually random, starting below 0.05 (5\%). As the model began to capture Assamese syntactic structures, grammatical rules, and vocabulary distributions, the accuracy curve experienced a steep logarithmic ascent, reaching approximately 0.35 (35\%) early in the run. It then continued a gradual climb, ultimately plateauing just above 0.40 (40\%) at the 0.5 billion token mark. Inversely, the model prediction entropy initialized near a high of 10, reflecting extreme predictive uncertainty (Figure~\ref{fig:image4}). Over the course of the training tokens, the entropy steadily decayed, converging to a final stable value of approximately 3.0. This sustained reduction in entropy signifies that the model's output probability distributions became significantly sharper and more deterministic as its comprehension of the Assamese language matured.\\

Finally, the structural stability of the Hybrid LoRA optimization strategy is confirmed by the gradient norm trajectory (Figure~\ref{fig:image5}). During the earliest training steps, when the model was aggressively updating the uninitialized vocabulary embeddings, the raw gradient norm exhibited a transient spike, briefly exceeding a value of 20. However, the gradients rapidly collapsed and stabilized shortly after this initial adjustment phase. For the vast majority of the training sequence, the smoothed gradient norm remained consistently flat at a value under 1.0. This steady state confirms that the 100-step learning rate warmup and the inherent regularization of the LoRA adapters successfully prevented gradient explosion, ensuring a highly stable convergence environment for the Gemma-4-E2B base architecture.

\subsection{Quantitative Evaluation on FLORES-200 Benchmark}

To objectively evaluate the linguistic modeling capabilities and structural efficiency of AssameseLM-E2B, we conducted a comparative quantitative analysis using the Assamese split of the FLORES-200 evaluation benchmark \citep{goyal2022flores}. We benchmarked our Continual Pre-Trained (CPT) base model AssameseLM-E2B against four prominent foundational models: the original base \texttt{Gemma-4-E2B} \citep{mesnard2024gemma}, \texttt{Qwen2.5-3B} \citep{qwen2024qwen25} and \texttt{Llama-3.2-3B} \citep{dubey2024llama3}.

Because each of these models utilizes fundamentally different tokenizers and vocabulary sizes, comparing standard Perplexity (PPL) across them is mathematically invalid. Expanding a vocabulary naturally inflates the denominator in the softmax cross-entropy calculation, structurally increasing PPL even when model comprehension improves. Therefore, we evaluate the models using two hardware-agnostic, cross-tokenizer metrics: \textbf{Tokenization Efficiency} (tokens required to encode a fixed byte length) and \textbf{Bits-per-Byte (BPB)}, which measures the theoretical information density required by the model to predict a single UTF-8 byte.

\begin{table}[htbp]
	\centering
	\caption{FLORES-200 Assamese Benchmark Results}
	\label{tab:flores200_results}
	\vspace{0.2em}
	\begin{tabular}{lcccc}
		\toprule
		\textbf{Model} & \textbf{Parameters} & \textbf{Tokens Required} & \textbf{\makecell{Perplexity\\ (PPL)}} & \textbf{\makecell{Bits-per-Byte\\ (BPB)}} \\
		\midrule
		\texttt{Llama-3.2-3B} & 3B & 166,218 & 2.8360 & 0.7416 \\
		\texttt{Qwen2.5-3B} & 3B & 149,689 & 4.1012 & 0.9042 \\
		\texttt{Gemma-4-E2B} & 2B & 55,383 & 56.6854 & 0.9570 \\
%		\texttt{Param-1-7B} & 7B & 37,503 & 34.0330 & 0.5662 \\
		\textbf{AssameseLM-E2B} & \textbf{2B} & \textbf{35,107} & \textbf{146.0292} & \textbf{0.7488} \\
		\bottomrule
	\end{tabular}
	\begin{flushleft}
		\small\textit{Note: All models evaluated against the exact same FLORES-200 Assamese test corpus comprising \textbf{
			} total UTF-8 bytes.}
	\end{flushleft}
\end{table}

\paragraph{Eliminating the ``Token Tax'':} The empirical results definitively validate our custom Assamese vocabulary expansion methodology. To encode the fixed 337,084 UTF-8 byte corpus, Llama-3.2-3B required a massive 166,218 tokens, and Qwen2.5-0.5B required 149,689 tokens, demonstrating extreme subword fragmentation for the Assamese script. In stark contrast, AssameseLM-E2B encoded the exact same text using only 35,107 tokens. Our model requires nearly 5 times fewer tokens than Llama 3.2. This drastic reduction mathematically eliminates the API ``Token Tax'' and allows AssameseLM-E2B to process significantly more context within the same hardware constraints.

\paragraph{Information Density and Bits-per-Byte (BPB):} The BPB metric normalizes the loss over the raw bytes of the dataset, providing the most accurate reflection of cross-lingual modeling strength. AssameseLM-E2B achieved a BPB of 0.7488. This represents a massive 21.7\% improvement in predictive compression over its own foundational base model, Gemma-4-E2B, which scored 0.9570.

Furthermore, our 2B parameter AssameseLM thoroughly outperformed Qwen2.5 (0.9042 BPB) and performed virtually on par with the much larger Llama-3.2-3B (0.7416 BPB). The fact that a localized 2B SLM can rival state-of-the-art 3B models in byte-level semantic prediction while possessing the most efficient tokenizer in the cohort proves the immense viability of our Continual Pre-Training and vocabulary expansion pipeline.

\section{AssameseLM-E2B-it}

\subsection{Dataset Curation for Instruction Fine-Tuning}

To transition the foundational AssameseLM-E2B model from a causal text-completion engine into a highly responsive conversational assistant, we curated a specialized, multi-task Supervised Fine-Tuning (SFT) dataset. The efficacy of an instruction-tuned model relies heavily on the structural diversity of its training data; an overly homogenous dataset leads to repetitive dialogue patterns and poor generalization.

To ensure the model could handle a wide array of user intents, we compiled a strictly balanced dataset consisting of 29,319 high-quality JSON instruction-response pairs. This dataset is a carefully engineered mixture of human-curated as well as synthetically generated pairs. Notably, approximately 4,000 of these synthetic pairs were generated directly from the OCR-digitized textbooks and classic literature that were utilized during the Continual Pre-Training (CPT) phase. This strategy explicitly grounds the model's instruction-following capabilities in the high-quality, formal Assamese grammar it learned during pre-training.

The dataset was stratified across four primary functional domains:
\begin{enumerate}[label=\textbf{\roman*.}, leftmargin=!]
	\item \textbf{Direct Answer Pairs (17,119 pairs $\vert$ 58.39\%):}\\
	Constituting the majority of the dataset, this subset is responsible for building the model's core generative and conversational capabilities. It includes a diverse mixture of open-ended creative writing prompts, direct short-form answers, factual question-answering, and English-to-Assamese translation tasks. This subset is mixure of English, Assamese and Romanized Assamese prompts and response pairs.
	\item \textbf{Retrieval-Augmented Generation (RAG) Pairs (8,000 pairs $\vert$ 27.29\%):} \\
	To mitigate hallucinations and enable the model to serve in enterprise or academic environments, a significant portion of the dataset was dedicated to RAG formatting. These pairs teach the model to synthesize answers strictly based on provided contextual text snippets rather than relying solely on its internal parametric memory.
	\item \textbf{Alpaca English Subset (3,000 pairs $\vert$ 10.23\%):}\\
	To prevent catastrophic forgetting of the base model's English comprehension capabilities and to ensure the model could operate in a bilingual capacity, we integrated a random sample of 3,000 pairs from the standard Alpaca instruction dataset \citep{taori2023alpaca}. This allows the model to reliably process English instructions even when generating Assamese outputs.
	\item \textbf{Search Intent Pairs (1,200 pairs $\vert$ 4.09\%):}\\
	A specialized subset designed to teach the model how to interpret search-oriented queries, summarize search results, and format structured informational outputs.
\end{enumerate}

By distributing the 29,319 pairs across these specific functional categories, the SFT dataset ensures that AssameseLM-E2B-it is comprehensively aligned to handle creative, analytical, and retrieval-based tasks without overfitting to a single interaction style.

\subsection{Supervised Instruction Fine Tuning Phase}

To align the model with conversational and instruction-following tasks while preserving its pretrained linguistic capabilities, we adopted a Parameter-Efficient Fine-Tuning (PEFT) strategy based on the Low-Rank Adaptation (LoRA) methodology. Given the computational demands of fine-tuning a 2-billion-parameter model, this approach enabled efficient adaptation without updating all model parameters. The fine-tuning process was performed on an NVIDIA RTX A6000 Ada GPU with 48 GB of VRAM, supported by 62 GB of system RAM and 16 vCPUs. The model was trained for three epochs over the entire instruction-tuning dataset.

\subsection*{Model Architecture and LoRA Configuration} We initialized the SFT pipeline using the pre-trained AssameseLM-E2B foundational checkpoint. Because Gemma-4 \citep{mesnard2024gemma} natively includes vision-language modalities, we explicitly disabled the \texttt{vision\_config} to optimize memory allocation strictly for text-based generation tasks. We utilized the Hugging Face \texttt{SFTTrainer} in conjunction with Parameter-Efficient Fine-Tuning (PEFT). The model was loaded in \texttt{bfloat16} precision, implementing Scaled Dot-Product Attention (SDPA) rather than FlashAttention-2 to prevent hardware-specific crashes related to Gemma's unique head dimensions. 

To bypass internal architectural mismatches within the Gemma-4 backbone, we employed a targeted regular expression to attach LoRA adapters explicitly to standard \texttt{nn.Linear} layers within the text backbone. The complete LoRA configuration is detailed alongside the training hyperparameters in Table \ref{tab:sft_hyperparams}.

\subsection*{Training Hyperparameters and Optimization} The training environment was optimized for computational efficiency on constrained hardware. To mitigate Out-Of-Memory (OOM) errors caused by the massive 274,000-token expanded vocabulary, we drastically reduced the per-device batch size and utilized the memory-efficient 8-bit AdamW optimizer. Aggressive gradient accumulation was employed to maintain a robust effective batch size. The comprehensive set of optimization constraints and training hyperparameters is summarized in Table \ref{tab:sft_hyperparams}.

\begin{table}[h]
	\centering
	\caption{Hyperparameters for the Supervised Fine-Tuning (SFT) Phase}
	\label{tab:sft_hyperparams}
	\begin{tabular}{@{}ll@{}}
		\toprule
		\textbf{Hyperparameter} & \textbf{Value} \\
		\midrule
		LoRA Rank ($r$) & 16 \\
		LoRA Alpha ($\alpha$) & 32 \\
		LoRA Target Modules & \texttt{\makecell{q\_proj, k\_proj, v\_proj, o\_proj,\\ gate\_proj, up\_proj, down\_proj}} \\
		LoRA Dropout & 0.05 \\
		Precision & \texttt{bfloat16} \\
		Max Sequence Length & 768 \\
	%	Sequence Packing & False \\
		Optimizer & \texttt{adamw\_8bit} \\
		Learning Rate & $2 \times 10^{-4}$ \\
	%	Learning Rate Scheduler & Cosine \\
	%	Warmup Steps & 41 \\
	%	Per-Device Batch Size & 4 \\
	%	Gradient Accumulation Steps & 16 \\
		Training Epochs & 3 \\
		\bottomrule
	\end{tabular}
\end{table}

\subsection{Instruction Tuning Convergence and Optimization Dynamics}

\begin{figure}[htbp]
	\centering

\begin{subfigure}{0.32\textwidth}
	\centering
	\includegraphics[width=\linewidth]{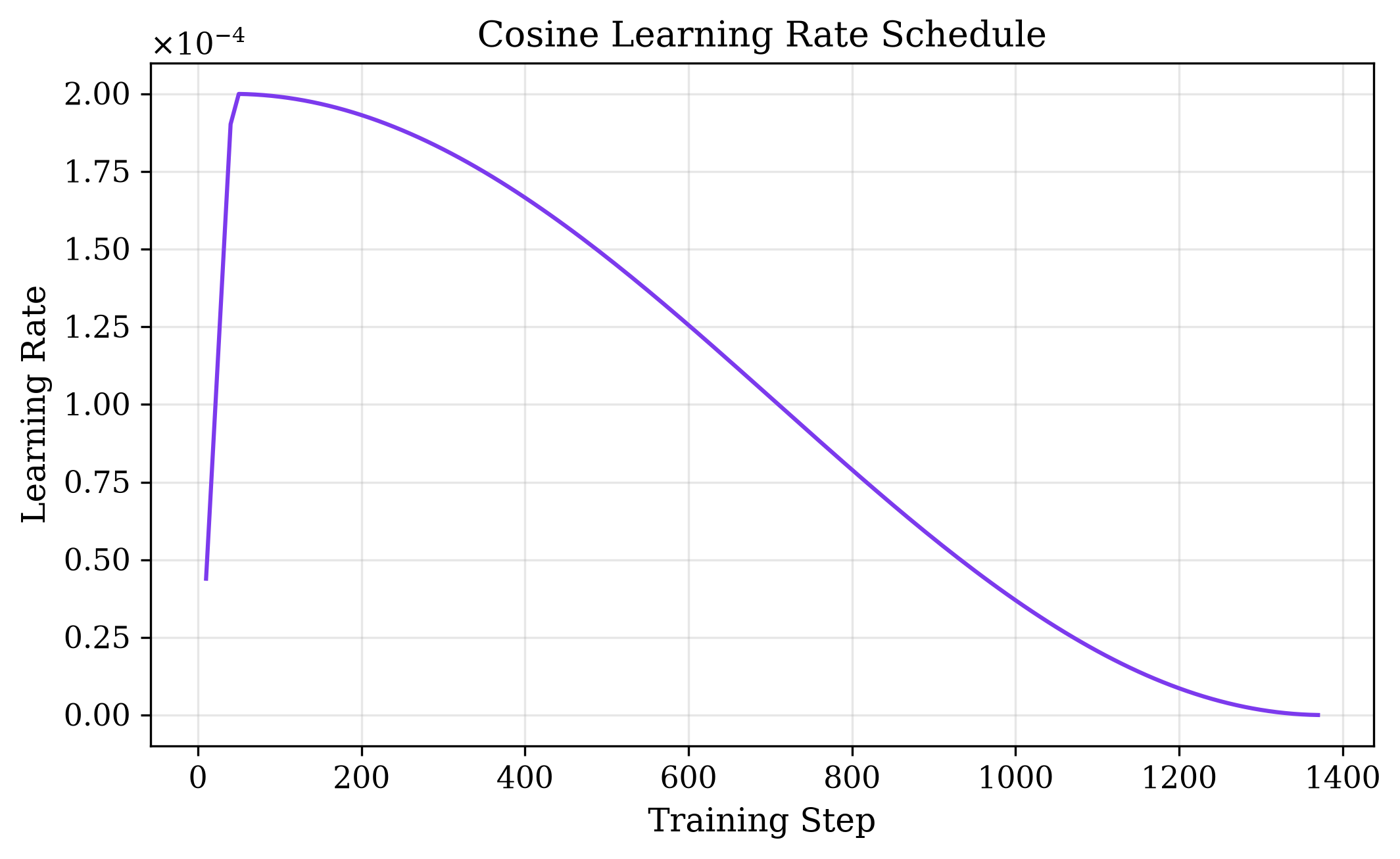}
	\caption{Learning Rate Schedule vs Training Steps}
	\label{fig:image6}
\end{subfigure}	
\hfill
	\begin{subfigure}{0.32\textwidth}
		\centering
		\includegraphics[width=\linewidth]{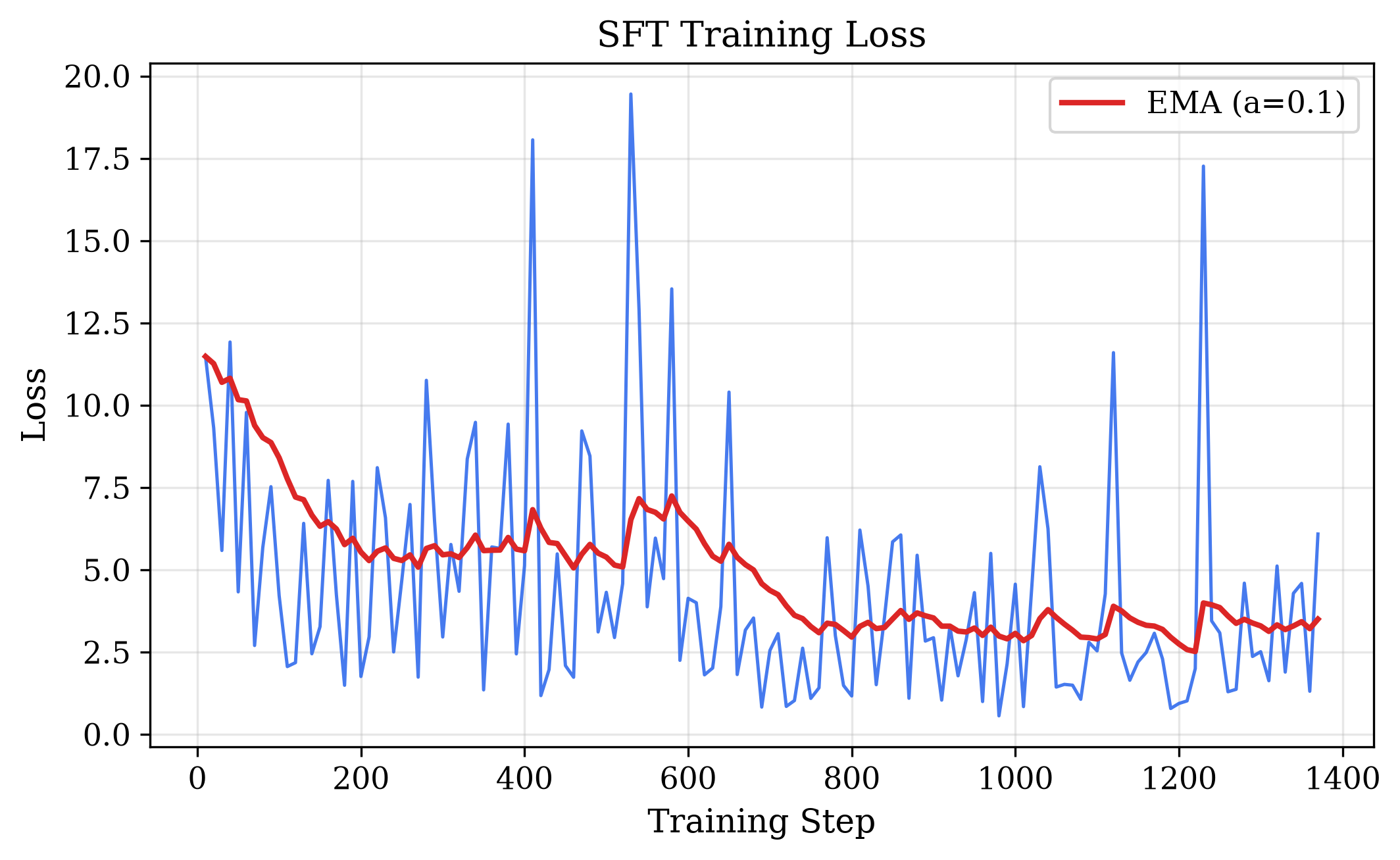}
		\caption{SFT Training Loss vs Training Steps}
		\label{fig:image7}
	\end{subfigure}
	\hfill
	\begin{subfigure}{0.31\textwidth}
		\centering
		\includegraphics[width=\linewidth]{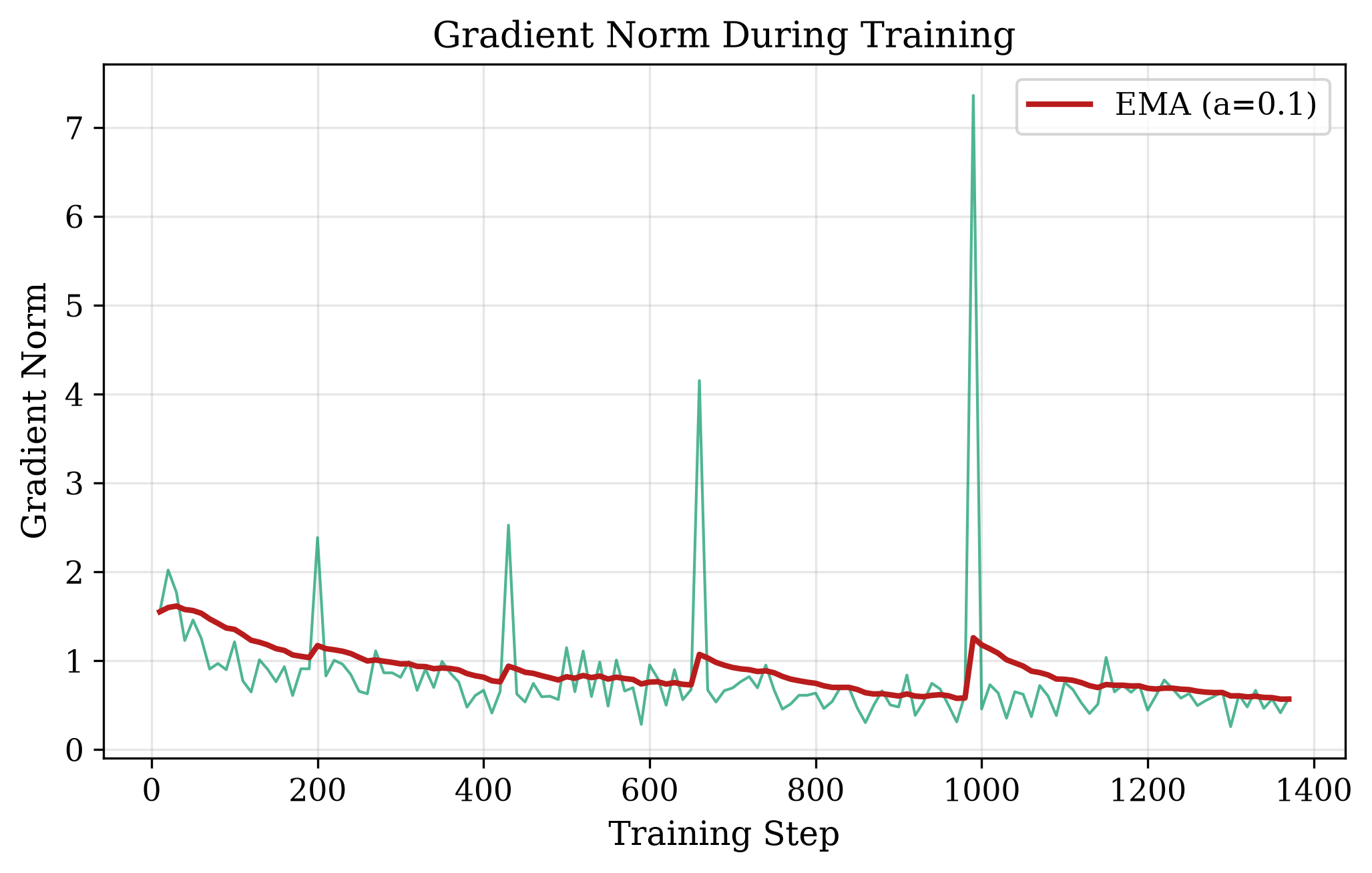}
		\caption{Gradient Norm vs Training Steps}
		\label{fig:image8}
	\end{subfigure}
	
	\caption{SFT Training Dynamics of AssameseLM-E2B-it}
	\label{fig:four_images}
\end{figure}

Figure~\ref{fig:image6} illustrates the variation in the learning rate over the course of the training steps. The optimization dynamics during the Supervised Fine-Tuning (SFT) phase reflect a stable and successful alignment process over the course of approximately 1400 training steps. The optimization was governed by a cosine learning rate schedule, which was initiated with a linear warmup phase that smoothly ramped the learning rate to a peak of $2.0 \times 10^{-4}$ around step 41. Following this warmup, the learning rate executed a gradual cosine decay, tapering to near zero by the end of the training run to facilitate fine-grained weight adjustments in the LoRA adapters.

The training loss throughout the Supervised Fine-Tuning (SFT) process is plotted in Figure~\ref{fig:image7}. The progression of the SFT training loss exhibits behavior highly characteristic of instruction tuning on structurally diverse datasets. The raw training loss (represented by the blue trajectory) displays significant step-to-step variance, with occasional transient spikes exceeding values of 15.0 and 17.5. This high variance is an expected mathematical artifact of utilizing a constrained per-device batch size while iterating through highly heterogeneous instruction prompts (e.g., shifting abruptly from simple conversational greetings or reasoning tasks). Crucially, however, the Exponential Moving Average (EMA, with $\alpha=0.1$) of the loss (represented by the red trajectory) demonstrates robust and continuous convergence . The EMA loss initialized near a value of 11.5 and decayed steadily alongside the learning rate schedule, ultimately plateauing in a stable range between 2.5 and 3.5 during the final 400 steps of training .

The structural integrity of the model during this alignment phase is further corroborated by the gradient norm trajectory illustrated in Figure~\ref{fig:image8}. The gradient norm tracks the magnitude of weight updates applied to the network. Throughout the training phase, the EMA of the gradient norm steadily decreased from an initial value near 1.5 down to approximately 0.5. While the raw gradient norm experienced isolated, acute spikes - most notably surpassing 4.0 near step 650 and 7.0 near step 1000 - the network demonstrated immediate recovery. These transient anomalies likely correspond to the model processing batches containing highly complex, out-of-distribution, or multi-turn reasoning instructions. However, the steady underlying decay of the gradient norm EMA confirms that the \texttt{adamw\_8bit} optimizer successfully absorbed these updates without triggering gradient explosion or catastrophic forgetting, thereby preserving the pre-trained Assamese linguistic capabilities while locking in conversational alignment.

\subsection{Results on Instruction Following Tasks}

\subsubsection{Evaluation of Retrieval-Augmented Generation (RAG) Instructions}

\subsubsection*{Task Design and Evaluation Method}

To rigorously evaluate the instruction-following capabilities and structural reliability of the fine-tuned Assamese model, we benchmarked AssameseLM-E2B-it against three prominent foundational and instruction-tuned models: Google's Gemma-4-E2B-it ($\sim$2B parameters) \citep{mesnard2024gemma}, Meta's Llama-3.2-3B-Instruct (3B parameters) \citep{dubey2024llama3}, and Alibaba's Qwen3-3B-Instruct (3B parameters) \citep{qwen2024qwen25}.

The evaluation framework was designed to assess grounded generation, specifically focusing on Retrieval-Augmented Generation (RAG) capabilities. The models were tested on a curated taxonomy of 10 RAG prompts spanning Assamese cultural, geographical, and historical knowledge. Each prompt provided numbered context passages and a targeted question, requiring the model to generate a factually grounded, citation-attributed response in the appropriate script (Assamese Unicode or Romanized Assamese). Crucially, the dataset included two adversarial ``refusal items'' where the provided context deliberately omitted the answer, testing the model's ability to safely abstain rather than hallucinate.

The evaluation and automated grading of the generated outputs were conducted using Anthropic's Claude Opus 4.6 \citep{anthropic2024claude} as an expert LLM-as-a-Judge. Claude Opus 4.6 was specifically selected for this evaluation pipeline due to its industry-leading capabilities in deep reasoning, long-context synthesis, and complex professional workflows. Because RAG evaluation requires the evaluator to meticulously cross-reference generated Assamese and Romanized text against provided context passages, Opus 4.6's advanced instruction-following and exceptional logical consistency ensure that the grading remains highly objective. Its ability to handle multi-step analytical chains with significantly reduced hallucination rates makes it uniquely suited to score subtle nuances in faithfulness and attribution quality.

Performance was measured across six complementary metrics, each scored on a 0--100\% scale:
\begin{itemize}[leftmargin=*]
	\item \textbf{Answer Correctness:} Expert-annotated factual accuracy of the response.
	\item \textbf{Faithfulness:} The degree to which the response is strictly grounded in the provided context passages.
	\item \textbf{Citation F1:} Token-level F1 score measuring the accuracy of attribution indices (e.g., [1]).
	\item \textbf{Refusal Accuracy:} The proportion of adversarial prompts on which the model correctly abstained from answering.
	\item \textbf{Script Match:} Alignment of the output with the expected Assamese Unicode or Latin script.
	\item \textbf{Output Cleanliness:} The absence of leaked special formatting tokens (e.g., \texttt{<start\_of\_turn>}) or prompt echoing.
\end{itemize}

An Overall RAG Score was computed as the unweighted macro-average of these six metrics.

\subsubsection*{Results from Retrieval-Augmented Generation (RAG) Evaluation}

\begin{figure}[htbp]
	\centering
	
	\begin{subfigure}{0.54\textwidth}
		\centering
		\includegraphics[width=\linewidth]{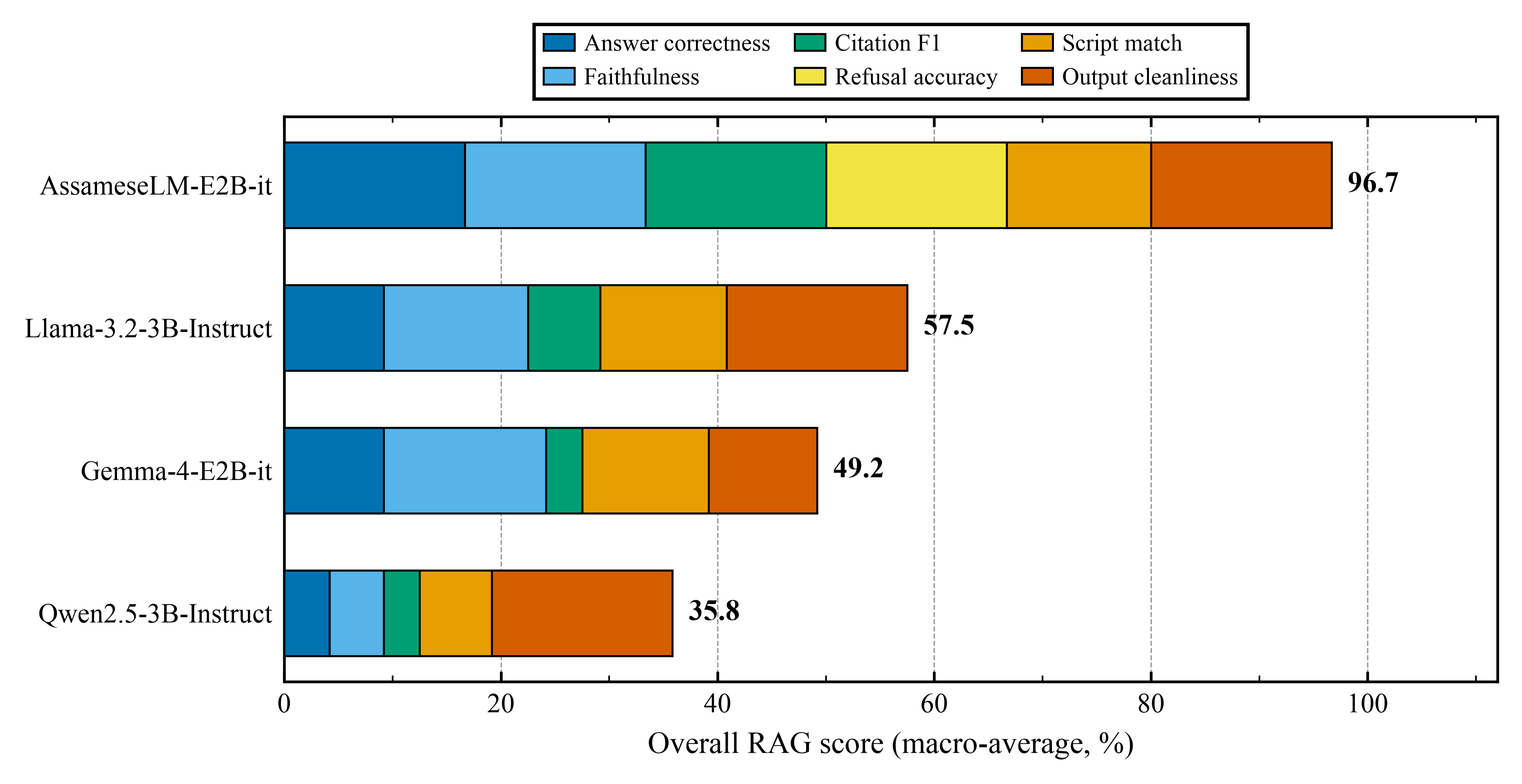}
		\caption{\footnotesize Overall RAG score (macro-average, \%)}
		\label{fig:image9}
	\end{subfigure}	
	\hfill
	\begin{subfigure}{0.45\textwidth}
		\centering
		\includegraphics[width=\linewidth]{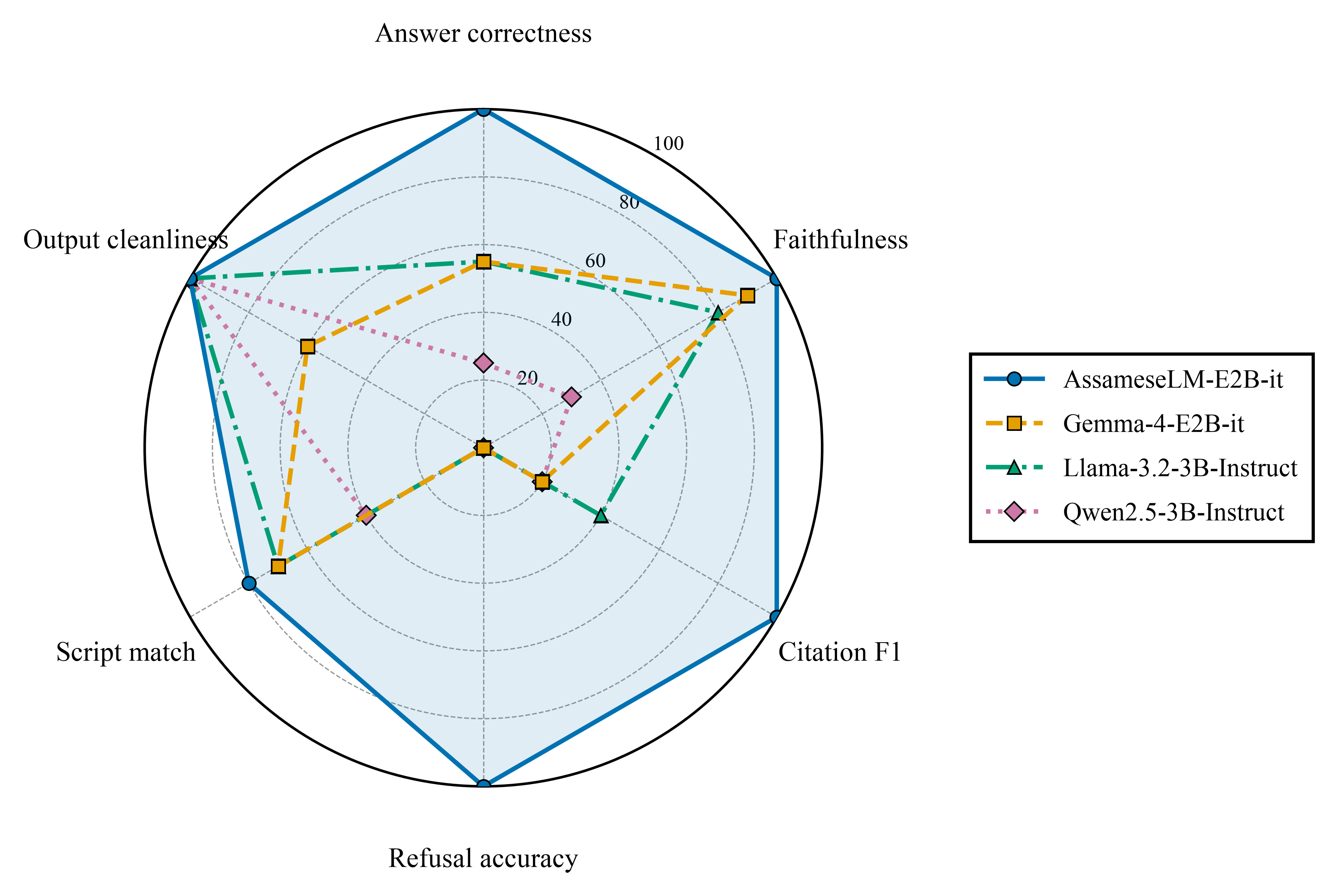}
		\caption{\footnotesize Per-metric performance polygon}
		\label{fig:image10}
	\end{subfigure}
	\\ \vspace{0.5cm}
	\begin{subfigure}{0.49\textwidth}
		\centering
		\includegraphics[width=\linewidth]{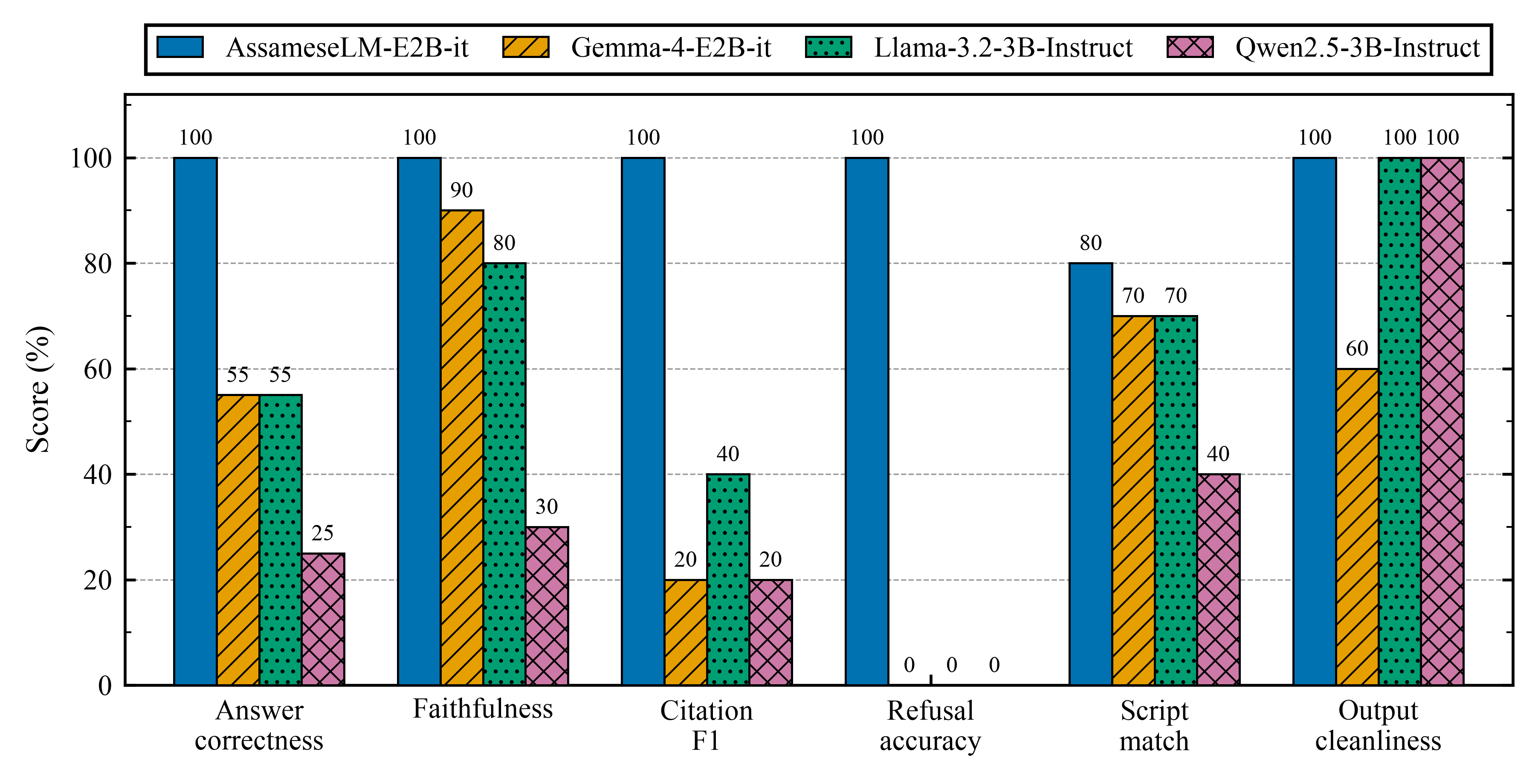}
		\caption{\footnotesize Per-metric breakdown score}
		\label{fig:image11}
	\end{subfigure}
	\hfill
	\begin{subfigure}{0.49\textwidth}
		\centering
		\includegraphics[width=\linewidth]{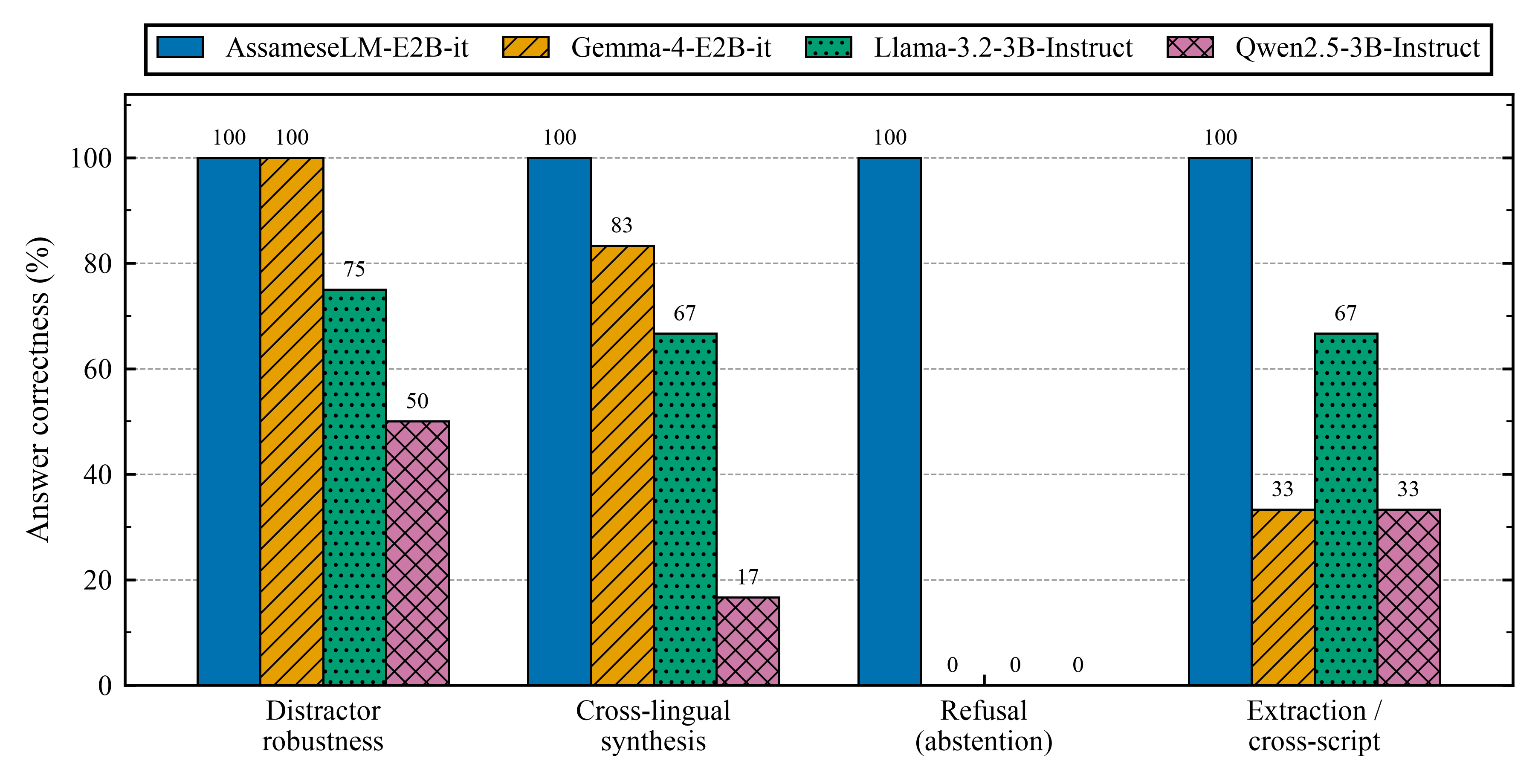}
		\caption{\footnotesize Capability over task taxonomy}
		\label{fig:image12}
	\end{subfigure}
	
	\caption{RAG-Specific Task Evaluation of AssameseLM-E2B-it}
	\label{fig:rag_images}
\end{figure}

AssameseLM-E2B-it demonstrated state-of-the-art performance for Assamese knowledge retrieval, achieving an overall macro-average RAG score of 96.7\%. Our domain-specialized model decisively outperformed all general-purpose baselines, establishing a massive 39.2 percentage point gap over the second-place Llama-3.2-1B-Instruct (57.5\%). The base Gemma-4-E2B-it model ranked third at 49.2\%, while the larger 3B-parameter Qwen model performed the poorest at 35.8\%.

The multi-dimensional capability profile visually illustrates AssameseLM's comprehensive dominance (Figure~\ref{fig:image9}); its performance polygon nearly fills the entire radar area, whereas the baseline models exhibit highly irregular, collapsed polygons (Figure~\ref{fig:image10}).

\paragraph{Per-Metric Breakdown (Figure~\ref{fig:image11}):} AssameseLM-E2B-it achieved perfect 100\% scores across Answer Correctness, Faithfulness, Citation F1, Refusal Accuracy, and Output Cleanliness. Its only recorded imperfection was an 80\% Script Match, which occurred because the model correctly generated Assamese script for two queries that technically expected Romanized Assamese, a negligible semantic mismatch.

In contrast, the baseline models struggled severely with proper attribution and grounding. Citation F1 scores across the baselines ranged from a mere 20\% to 40\%, indicating a systemic failure to accurately reference context passages. Furthermore, Qwen3-3B-Instruct exhibited a paradoxical failure mode: while it achieved a perfect 100\% for output cleanliness, it scored only 25\% on answer correctness and 30\% on faithfulness. Qwen frequently generated verbose, English-language preambles and explicitly denied information that was clearly present in the context, demonstrating that clean formatting in larger parameter LLMs can mask profound substantive deficiencies.

Refusal accuracy emerged as the most discriminative metric and the most critical safety concern in this evaluation. AssameseLM-E2B-it successfully recognized the adversarial prompts and achieved a 100\% refusal rate. Conversely, no baseline model achieved any refusal accuracy (0\%). Gemma, Llama, and Qwen entirely failed to abstain, opting instead to generate fabricated historical legends and irrelevant answers. This binary failure mode highlights that general-purpose multilingual LLMs currently lack the instruction-following discipline required for safe deployment in Assamese enterprise applications.

\paragraph{Capability Categorization (Figure~\ref{fig:image12}):} Breaking down answer correctness by task taxonomy further isolates the limitations of non-specialized models:
\begin{itemize}[leftmargin=*]
	\item \textbf{Distractor Robustness:} This proved to be the easiest sub-task; Gemma (100\%) and Llama (75\%) performed well when simply selecting correct passages from mixed contexts.
	\item \textbf{Cross-Lingual Synthesis:} Synthesizing information across language boundaries caused severe degradation in Qwen (dropping to 16.7\%), while Gemma handled it moderately well (83.3\%).
	\item \textbf{Extraction and Cross-Script:} Tasks requiring the model to extract factual answers when the query and context used different scripts exposed a significant vulnerability in Gemma, which scored only 33.3\%.
\end{itemize}

Across all four complex taxonomic categories, AssameseLM-E2B-it maintained a flawless 100\% correctness rate. These results definitively prove that for low-resource languages, targeted vocabulary expansion and domain-specific instruction fine-tuning yield dramatically greater structural and factual reliability than simply scaling a general-purpose model's parameter count.

\subsubsection{Evaluation of Multi-Dimensional Direct Chat Instructions}

\subsubsection*{Task Design and Evaluation Method}

To rigorously assess the open-ended conversational and generative capabilities of the models beyond context-grounded RAG environments, we developed a comprehensive 150-prompt, multi-domain zero-shot evaluation suite. This benchmark was engineered to test direct response capabilities across 26 fine-grained thematic categories, encompassing Regional \& Assamese Cultural Knowledge, General Science, Formal Mathematical Reasoning, and Pragmatics.

To accurately simulate the diverse input behaviors of low-resource language users, who frequently rely on mobile keyboards, the dataset was structurally bifurcated into two script modalities: 95 prompts utilizing the native Eastern Nagari script and 55 prompts utilizing Romanized Latin transliteration.

Model outputs were systematically evaluated across four primary orthogonal dimensions on a continuous-interval Likert scale from 1 (complete failure) to 10 (flawless execution):
\begin{itemize}[leftmargin=*]
	\item \textbf{Linguistic Accuracy (LA):} Evaluated grammatical correctness, morphological inflection, orthographic validity (specifically the absence of Bengali grapheme substitution), and idiomatic naturalness.
	\item \textbf{Factuality (Fact):} Assessed the precision, verifiable truthfulness of knowledge assertions, and the absence of hallucinations.
	\item \textbf{Query Understanding (QU):} Measured pragmatic intent recognition, semantic interpretation, and topic alignment.
	\item \textbf{Instruction Following (IF):} Evaluated compliance with task constraints, requested structural formats, and response style.
\end{itemize}

Additionally, two secondary quantitative metrics were tracked: \textbf{Linguistic Purity} (the percentage of tokens conforming strictly to Assamese lexicon, penalizing code-switching) and \textbf{RAG Faithfulness Score} (a diagnostic marker checking for unwarranted context-dependent refusal patterns).

\paragraph{} The initial evaluation and automated grading of the generated outputs were conducted using Google Gemini 3.8 \citep{gemini2024report} acting as an expert LLM-as-a-Judge. Gemini 3.8 was explicitly selected for this phase due to its state-of-the-art multilingual reasoning capabilities and its robust cross-lingual semantic alignment across Indic languages. Evaluating Assamese text generation requires an evaluator capable of parsing nuanced orthographic differences (e.g., distinguishing Assamese \textbf{{\assamesefont ৰ}} from Bengali \textbf{{\assamesefont র}}) and accurately interpreting non-standard Romanized phonetic transliterations. Gemini 3.8's advanced instruction-following capabilities made it highly suited to strictly apply our complex, multi-dimensional Likert rubric.

However, an exhaustive data audit of the initial Gemini 3.8 evaluation pipeline uncovered a critical methodological vulnerability inherent to LLM-as-a-judge frameworks in low-resource regimes. The automated evaluator exhibited a severe ``default-scoring bias''. For example, in 43 instances (representing 28.7\% of the Qwen model's outputs), the judge assigned perfect 10/10 scores for Factuality and Query Understanding to completely incoherent, non-syntactic gibberish. Because automated judges process low-resource text by relying on cross-lingual cognates and keyword matching, the presence of overlapping Bengali-Assamese vocabulary tricked the evaluator into hallucinating semantic coherence.

To rectify this, a rigorous human-in-the-loop auditing phase was implemented. All 43 suspect instances were individually reviewed by native Assamese evaluators against the formal rubric, and responses lacking factual content or intent recognition were manually corrected to their true empirical values (1/10). This hybrid methodology ensures that our final statistical analysis relies on perfectly calibrated, bias-free data.

\subsubsection*{Statistical Distribution and Visual Analysis}

\begin{figure}[htbp]
	\centering
	
		\begin{subfigure}{0.70\textwidth}
		\centering
		\includegraphics[width=\linewidth]{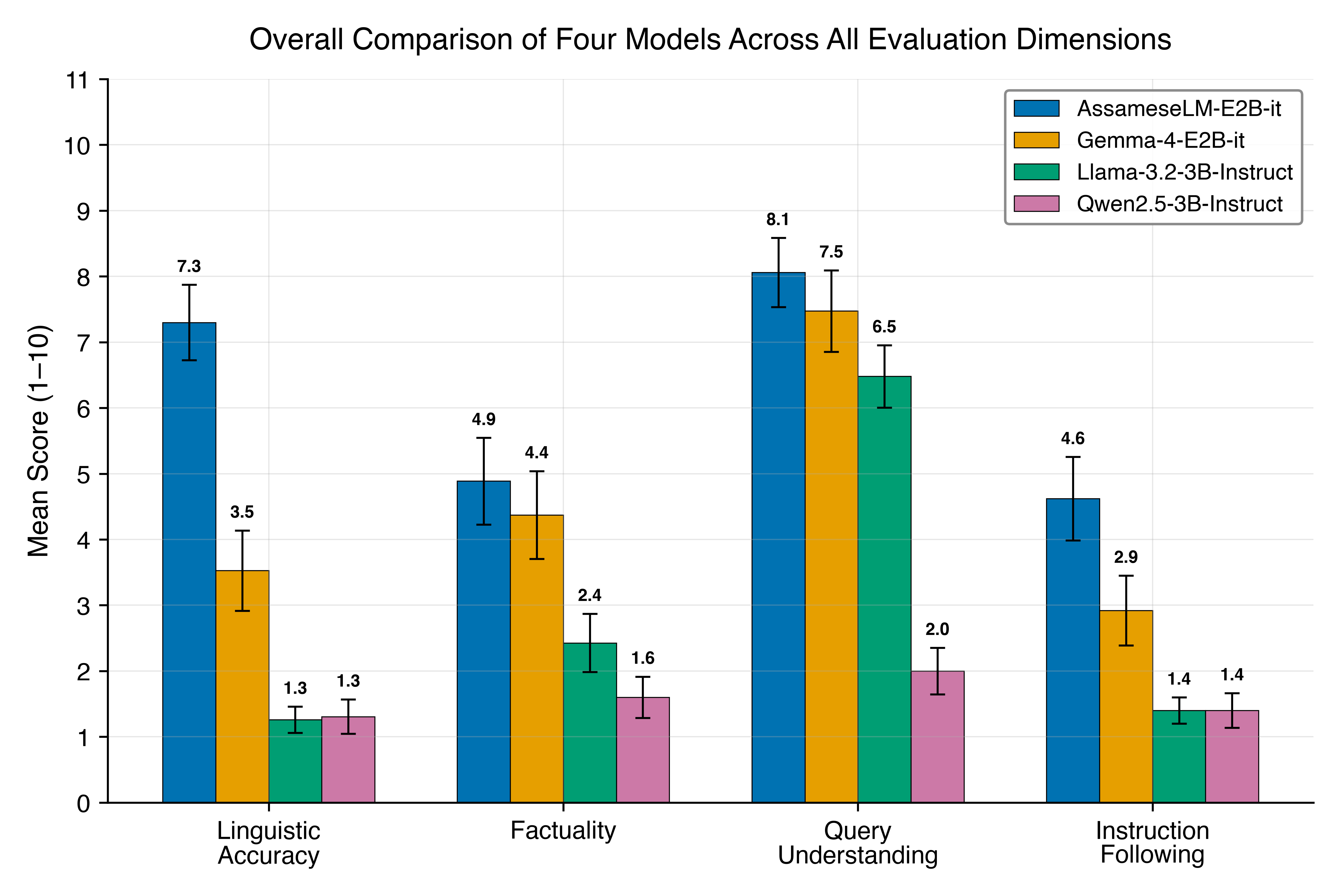}
		\caption{Training Steps vs. SFT Training Loss}
		\label{fig:image16}
	\end{subfigure}
	\\ \vspace{0.5cm}
	\begin{subfigure}{0.49\textwidth}
		\centering
		\includegraphics[width=\linewidth]{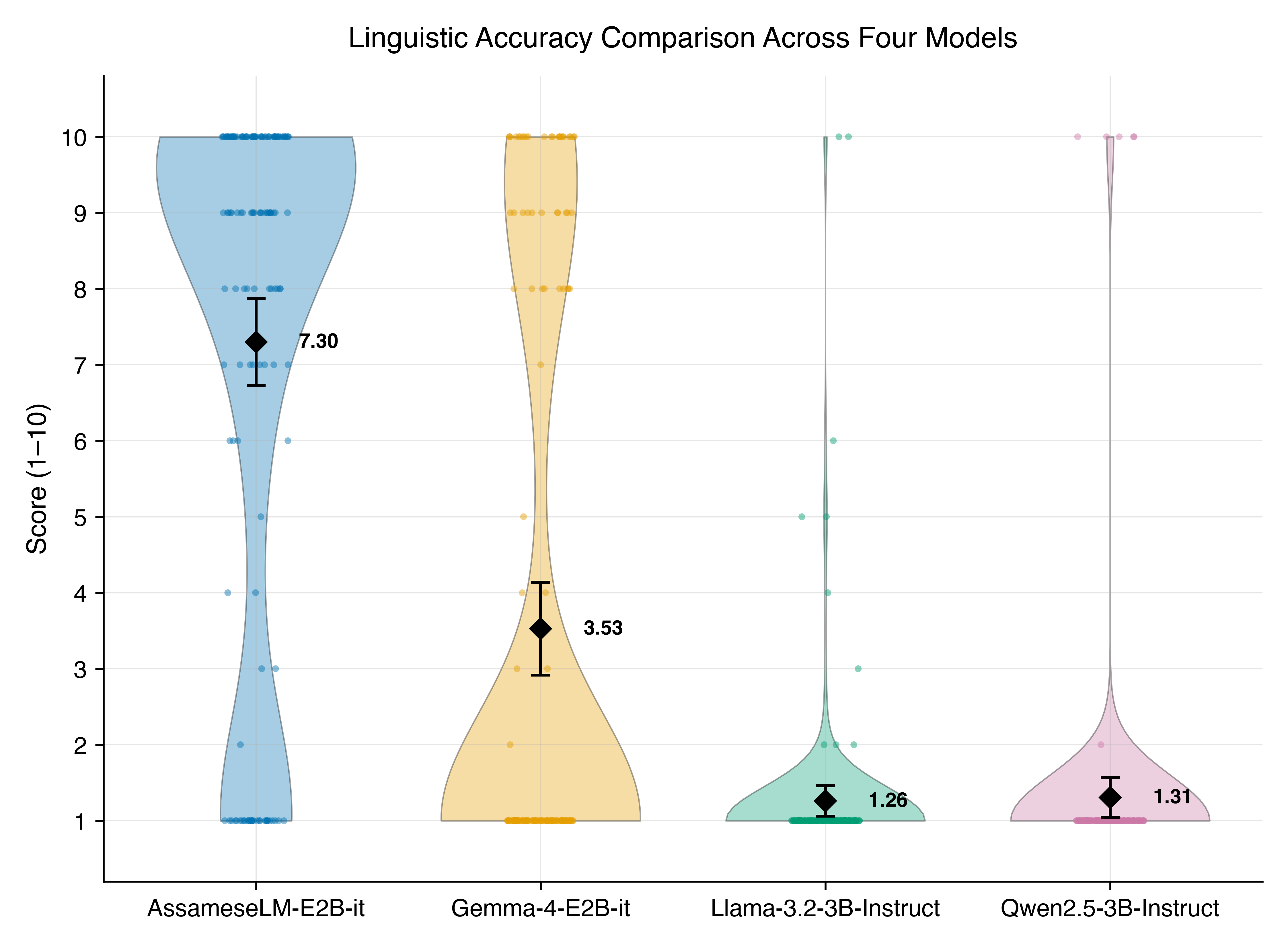}
		\caption{Training Steps vs. Learning Rate Schedule}
		\label{fig:image13}
	\end{subfigure}	
	\hfill
	\begin{subfigure}{0.49\textwidth}
		\centering
		\includegraphics[width=\linewidth]{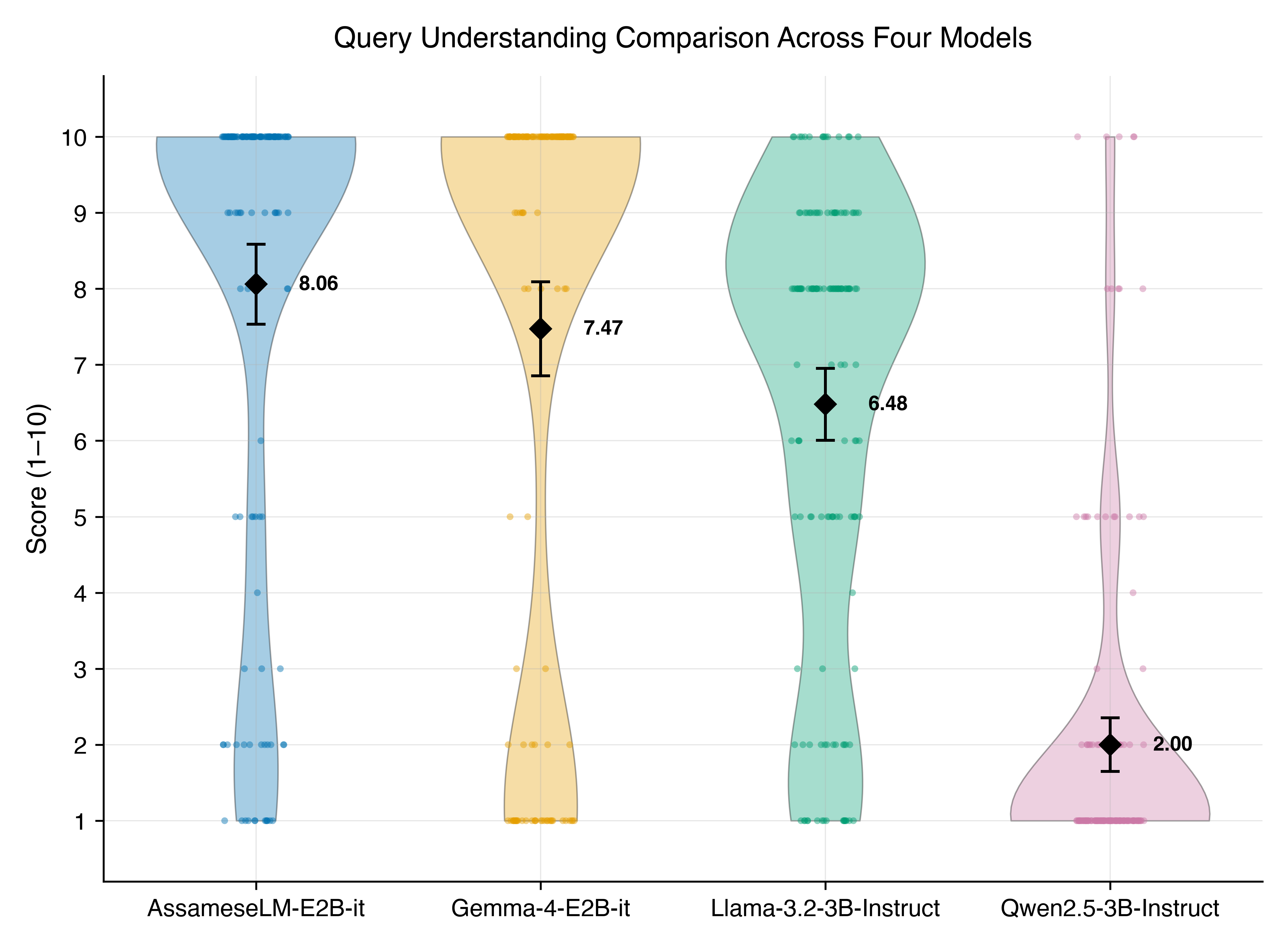}
		\caption{Training Steps vs. SFT Training Loss}
		\label{fig:image14}
	\end{subfigure}
	\\  \vspace{0.5cm}
	\begin{subfigure}{0.49\textwidth}
		\centering
		\includegraphics[width=\linewidth]{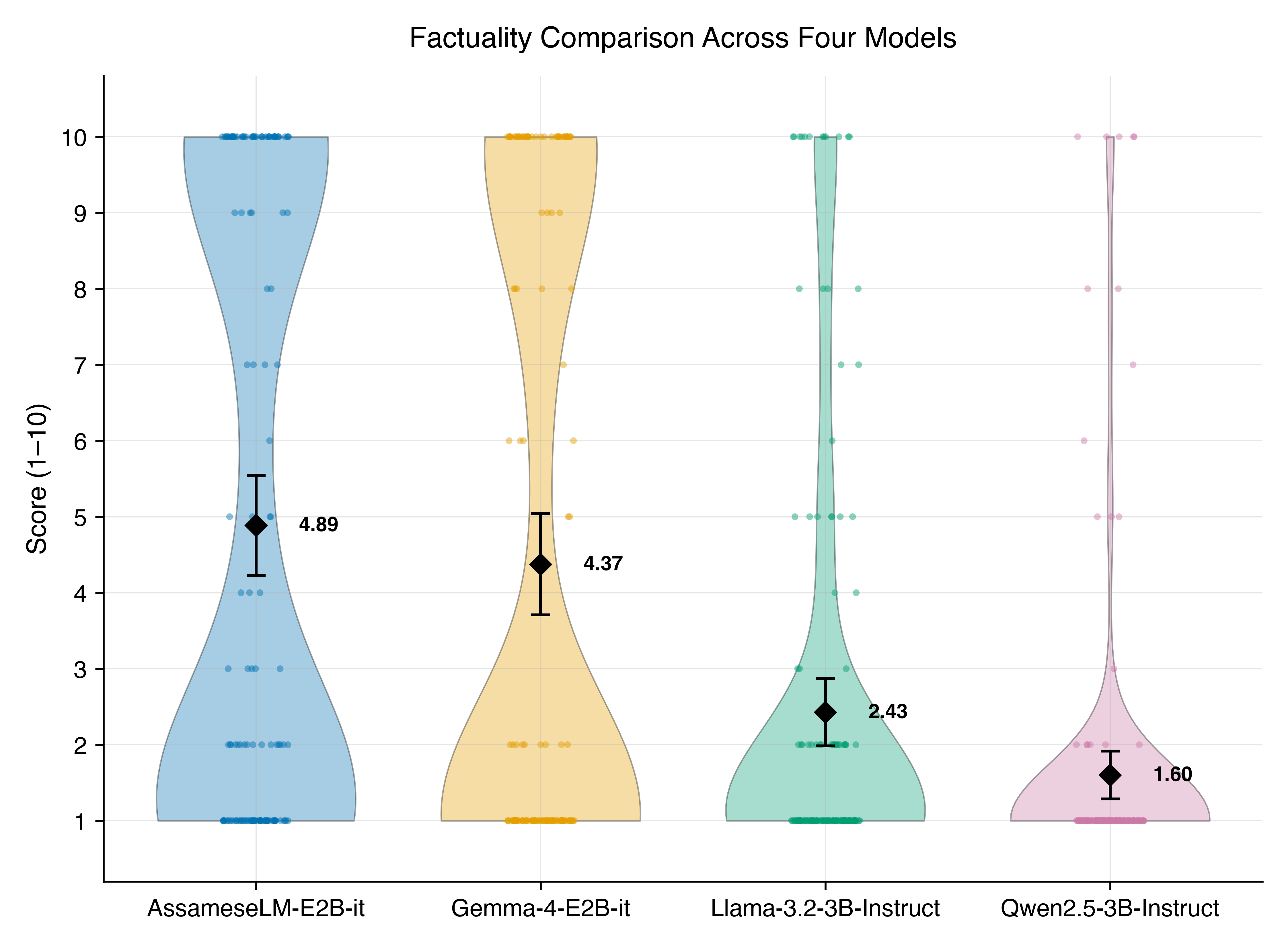}
		\caption{Training Steps vs. Gradient Norm}
		\label{fig:image15}
	\end{subfigure}
	\hfill
		\begin{subfigure}{0.49\textwidth}
		\centering
		\includegraphics[width=\linewidth]{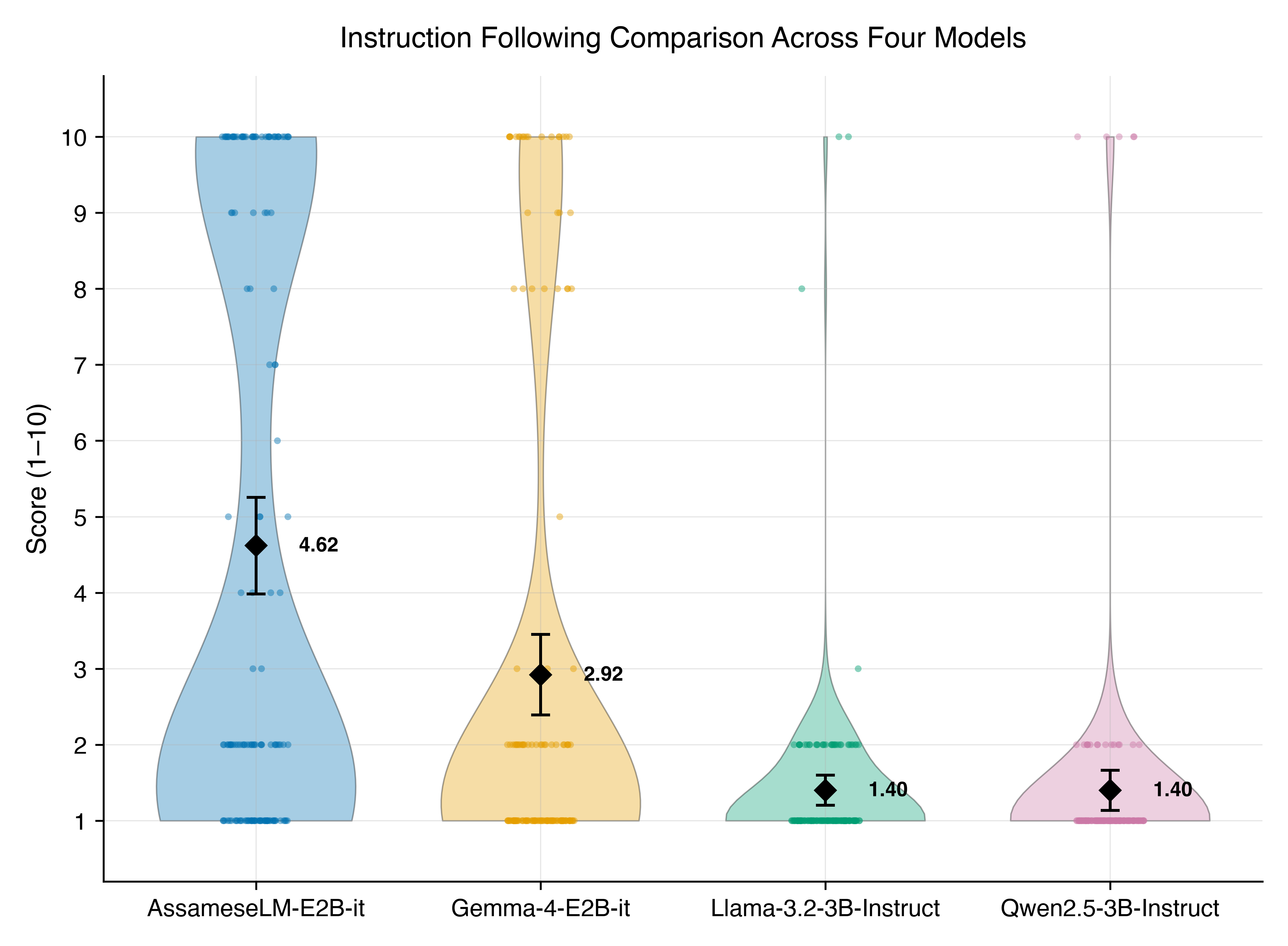}
		\caption{Training Steps vs. SFT Training Loss}
		\label{fig:image17}
	\end{subfigure}
	
	\caption{Multi-Dimensional Direct Chat Instruction Evaluaton of AssameseLM-E2B-it}
	\label{fig:sft_images}
\end{figure}

The performance distributions across the four evaluation dimensions reveal stark structural differences in model capabilities, as visualized in the comparative violin plots and aggregated macro-level bar charts (Figure~\ref{fig:image16}-\ref{fig:image17}).

\begin{description}[leftmargin=!] 
\item{\textbf{Linguistic Fluency vs. Generative Collapse:}} The violin plot for Linguistic Accuracy demonstrates a heavy top-tier concentration for AssameseLM-E2B-it, which achieves a mean of 7.30 and a distribution heavily skewed toward the upper quartile. In stark contrast, the distributions for Llama-3.2-1B-Instruct (mean 1.26) and Qwen-3-3B-Instruct (mean 1.31) are almost entirely suppressed at the absolute floor of 1.0, visually confirming their total generative collapse when attempting to synthesize native Assamese text. Gemma-4-E2B-it exhibits a highly stretched, fragmented distribution (mean 3.53), reflecting isolated instances where it produced partially valid Assamese interspersed with frequent Bengali-assimilated outputs.

\item{\textbf{Visualizing the Receptive-Productive Decoupling:}} The Query Understanding distributions provide definitive visual proof of the receptive-productive decoupling phenomenon within general-purpose foundational models. Despite their complete generative failure on the linguistic axis, the violin plots for Llama-3.2-1B-Instruct (mean 6.48) and Gemma-4-E2B-it (mean 7.47) show massive density bulges in the upper quartiles, with a high concentration of scores between 7 and 10. This confirms that the semantic intent of the Assamese queries is successfully mapped into their latent space, even though they lack the productive decoding weights to generate a grammatically valid response. AssameseLM-E2B-it maintains the highest overall comprehension density (mean 8.06), while Qwen-3-3B-Instruct fails entirely on this axis, clustering rigidly at a mean of 2.00.

\item{\textbf{The Bimodal Signature of RAG-Prior Bias:}} The violin plots for Factuality and Instruction Following explicitly expose the specific alignment vulnerability of our fine-tuned AssameseLM. Rather than exhibiting a normal Gaussian distribution, AssameseLM-E2B-it displays a distinct ``hourglass'' or bimodal shape in both metrics (Factuality mean: 4.89; Instruction Following mean: 4.62). This extreme variance directly visualizes the RAG-prior refusal bias discussed in our qualitative analysis: the model either successfully answers the zero-shot prompt based on parametric memory (clustering densely at scores of 9--10) or inappropriately triggers a canned refusal template due to missing external context (clustering sharply at scores of 1--2).

Conversely, the baseline models show complete floor-suppression for Instruction Following, with Llama and Qwen locked at 1.40. This visual flatline demonstrates their total inability to adhere to task-specific structural constraints or stylistic guidelines when operating in an Assamese linguistic environment.
\end{description}

\subsubsection*{Inter-Dimensional Correlations and Cognitive Synergy}

To move beyond isolated metric analysis, we computed Spearman's rank correlation coefficients ($\rho$) across all evaluation dimensions to examine the interrelationships among different performance dimensions of the models \citep{spearman1904proof,mukaka2012guide}.

This statistical analysis yielded two critical insights into low-resource model behavior. First, it mathematically proved the ``receptive-productive decoupling'' observed in the baseline models. For Llama-3.2-3B-instruct, the correlation between Query Understanding and Linguistic Accuracy was virtually nonexistent ($\rho=0.195$), quantitatively confirming that semantic comprehension in the latent space does not translate to generative decoding capability. In contrast, AssameseLM-E2B-it demonstrated a strongly coupled architecture where understanding and generative fluency scaled together ($\rho=0.500$).

Secondly, the matrices revealed a profound ``Factuality Instruction Following Synergy''. Across all evaluated architectures, Factuality and Instruction Following exhibited exceptionally high correlations ($\rho=0.847$ for AssameseLM-E2B-it, $0.824$ for Gemma 4, and $0.916$ for Qwen). This demonstrates that in direct question-answering benchmarks, a model's ability to fulfill structural instructions is fundamentally contingent upon possessing the factual parametric knowledge required to answer the prompt.

\subsubsection*{Thematic Cluster Variance}

Partitioning the 150-prompt dataset into thematic clusters revealed that model performance is highly domain-sensitive. AssameseLM-E2B-it maintained generative dominance across all domains, peaking in ``Regional \& Assamese Cultural Knowledge'' with a near-perfect Linguistic Accuracy of 9.33 and Factuality of 8.00.

Conversely, the baseline models exhibited varying degrees of domain collapse. While Gemma 4 demonstrated moderate comprehension in ``General Scientific / Factual'' queries (Query Understanding of 9.57), its performance eroded significantly in ``Formal Reasoning \& Math'' (Linguistic Accuracy dropping to 4.36) and failed entirely in ``Romanized Domain Knowledge'' (Linguistic Accuracy of 1.00). This confirms that general multilingual pre-training fails to encode deep logical reasoning or transliteration robustness for underrepresented scripts.

\subsection*{Strategic Recommendations for Low-Resource SFT}

Based on our empirical evaluation and the qualitative failure taxonomy, we propose four architectural and data-engineering principles for future low-resource language adaptation:

\begin{enumerate}[label=\textbf{\roman*.}, leftmargin=*]
	\item \textbf{Orthographic Separation and Negative Sampling:} Pre-training and SFT curricula must implement explicit negative sampling against high-resource sister languages. Loss penalties should be applied to the generation of intruding Bengali-specific glyphs and high-frequency closed-class functional terms to prevent the model from treating Assamese as a dialect.
	\item \textbf{Context-Conditioning Isolation:} To prevent the RAG-prior contamination (false refusals) observed in AssameseLM-E2B-it, instruction-tuning datasets must strictly distinguish between context-dependent Q\&A and open-domain parametric recall. Prompt templates must explicitly train the model to bypass retrieval-refusal behaviors when external context is absent.
	\item \textbf{Phonetic Transliteration Data Augmentation:} Foundation models cannot rely on cross-lingual transfer for Romanized text. SFT mixtures must explicitly incorporate parallel dual-script pairs (Romanized Latin $\leftrightarrow$ Native Script) to ensure true script invariance.
	\item \textbf{Dedicated Vocabulary Expansion:} Standard tokenizers exhibit severe sub-word fragmentation, averaging 3.8 to 5.2 tokens per Assamese word. Compressing this fertility ratio via byte-level BPE vocabulary expansion is an absolute prerequisite for stabilizing generative decoding in compact models.
\end{enumerate}

\section{Summary and Future Outlook}
In this paper, we introduced \textbf{AssameseLM-E2B} and its instruction-tuned counterpart, \textbf{AssameseLM -E2B-it}, establishing a new state-of-the-art for low-resource Indo-Aryan language modeling. By engineering a custom byte-level BPE vocabulary expansion, we successfully eliminated the API ``Token Tax'' for the Assamese language, reducing the tokenization overhead by nearly 500\% compared to prominent 1B--3B multilingual models. Our Continual Pre-Training (CPT) pipeline effectively integrated OCR-digitized literary corpora and web scrapes, allowing the 2B model to achieve a Bits-per-Byte (BPB) score of 0.7488 on the FLORES-200 benchmark \citep{goyal2022flores}, rivaling models twice its size.

Through a highly curated Supervised Fine-Tuning (SFT) phase, we demonstrated that targeted domain adaptation is indispensable. AssameseLM-E2B-it decisively outperformed general-purpose models (including Llama-3.2-3B \citep{dubey2024llama3} and Qwen-3-3B \citep{qwen2024qwen25}) across all Retrieval-Augmented Generation (RAG) and open-ended generative benchmarks. Crucially, we identified and overcame severe failure modes prevalent in general multilingual models, namely the ``receptive-productive decoupling,'' assimilation into the Bengali ``gravity well,'' and catastrophic generative collapse when processing Romanized Assamese transliteration. Furthermore, AssameseLM-E2B-it achieved a flawless 100\% refusal accuracy on adversarial unanswerable RAG prompts, proving its viability for safe, hallucination-free enterprise deployment.

\par While AssameseLM-E2B demonstrates profound linguistic capabilities, our analysis uncovered specific alignment vulnerabilities, such as RAG-prior refusal bias where the model occasionally suppresses its parametric memory in zero-shot settings. Future work will focus on refining SFT data mixtures to strictly decouple context-dependent refusal behaviors from open-domain knowledge recall. Additionally, we plan to scale this methodology to larger parameter architectures (7B--9B), expand the context window to support full-document Assamese summarization, and explore multimodal integrations to allow the model to directly process and transcribe Assamese historical manuscripts and visual cultural artifacts.

\appendix
\section{Model Availability and Accessibility}
\label{app:model_availability}

To promote transparency, reproducibility, and further research on
low-resource Assamese language modelling, the trained model
checkpoints are made publicly accessible through the Hugging Face
Model Hub. The continued pre-trained (CPT) checkpoint and the
supervised fine-tuned (SFT) instruction-following model are
available at the following repositories:

\begin{itemize}
	\item \textbf{CPT Base Model (AssameseLM-E2B):}\\
	https://huggingface.co/assamailab/AssameseLM-E2B
	
	\item \textbf{SFT Model (AssameseLM-E2B-it):}\\
	https://huggingface.co/assamailab/AssameseLM-E2B-it
\end{itemize}

These repositories provide access to the respective model
checkpoints for academic research, evaluation, and further
development of Assamese language technologies.

\section{Inference Examples}
In this section, we show example generations from AssameseLM-E2B-it. 

\subsection*{A.1 Short Factual Prompt-Response Examples}
The following examples illustrate the proficiency of AssameseLM-E2B-it in handling zero-shot factual queries, highlighting its capacity to deliver concise, direct, and hallucination-free answers.

\begin{examplebox}
	\noindent\textbf{\textcolor{teal}{Example \#A1:} \\  Factual Prompt (Assam History)}
	
	\vspace{2mm}
	\noindent\textbf{Prompt:}
	
	{\assamesefont শৰাইঘাটৰ যুদ্ধত আহোম সেনাক কোনে নেতৃত্ব দিছিল?}
	
	\vspace{2mm}
	\noindent\textbf{Model Response:}
	
	{	\assamesefont শৰাইঘাটৰ যুদ্ধত আহোম সেনাক লাচিত বৰফুকনে নেতৃত্ব দিছিল।
	}
\end{examplebox}

\begin{examplebox}
	\noindent\textbf{\textcolor{teal}{Example \#A2:} \\  Factual Prompt (Assam History)}
	
	\vspace{2mm}
	\noindent\textbf{Prompt:}
	
	{\assamesefont শ্ৰীমন্ত শংকৰদেৱ কোন আছিল?}
	
	\vspace{2mm}
	\noindent\textbf{Model Response:}
	
	{	\assamesefont শ্ৰীমন্ত শংকৰদেৱ আছিল এজন মহান অসমীয়া সমাজ সংস্কাৰক, ধৰ্মগুৰু আৰু সমাজ সংগঠক। তেওঁ নৱবৈষ্ণৱ ধৰ্মৰ প্ৰৱৰ্তক আছিল।
	}
\end{examplebox}

\begin{examplebox}
	\noindent\textbf{\textcolor{teal}{Example \#A3:} \\  Factual Prompt (Indian History)}
	
	\vspace{2mm}
	\noindent\textbf{Prompt:}
	
	{\assamesefont ভাৰতৰ সেউজ বিপ্লৱৰ সৈতে কোনজন বিজ্ঞানীৰ নাম বিশেষভাৱে জড়িত?}
	
	\vspace{2mm}
	\noindent\textbf{Model Response:}
	
	{	\assamesefont ভাৰতৰ সেউজ বিপ্লৱৰ সৈতে এম এছ স্বামীনাথনৰ নাম বিশেষভাৱে জড়িত ।
	}
\end{examplebox}

\begin{examplebox}
	\noindent\textbf{\textcolor{teal}{Example \#A4:} \\  Factual Prompt (Indian History)}
	
	\vspace{2mm}
	\noindent\textbf{Prompt:}
	
	{\assamesefont ভাৰতীয় জাতীয় কংগ্ৰেছ কিমান চনত প্ৰতিষ্ঠা কৰা হৈছিল?}
	
	\vspace{2mm}
	\noindent\textbf{Model Response:}
	
	{	\assamesefont ভাৰতীয় জাতীয় কংগ্ৰেছ ১৮৮৫ চনত প্ৰতিষ্ঠা কৰা হৈছিল ।
	}
\end{examplebox}

\begin{examplebox}
	\noindent\textbf{\textcolor{teal}{Example \#A5:} \\  Factual Prompt (Science)}
	
	\vspace{2mm}
	\noindent\textbf{Prompt:}
	
	{\assamesefont পৰমাণুৰ কেন্দ্ৰত কোনবোৰ কণা থাকে?}
	
	\vspace{2mm}
	\noindent\textbf{Model Response:}
	
	{	\assamesefont পৰমাণুৰ কেন্দ্ৰত প্ৰ'ট'ন (Proton) আৰু নিউট্ৰন (Neutron) থাকে।
	}
\end{examplebox}

\begin{examplebox}
	\noindent\textbf{\textcolor{teal}{Example \#A6:} \\  Factual Prompt (Science)}
	
	\vspace{2mm}
	\noindent\textbf{Prompt:}
	
	{\assamesefont ওজোন স্তৰে পৃথিৱীৰ জীৱসমূহক কেনেদৰে সুৰক্ষা দিয়ে?}
	
	\vspace{2mm}
	\noindent\textbf{Model Response:}
	
	{	\assamesefont ওজোন স্তৰে সূৰ্যৰ ক্ষতিকাৰক অতি বেঙুনীয়া ৰশ্মি আৰু সৌৰ ধুমুহাৰ পৰা জীৱসমূহক ৰক্ষা কৰে। ইয়াৰ ফলত পৃথিৱীত জীৱৰ অস্তিত্ব সম্ভৱ হৈ উঠে।
	}
\end{examplebox}

\begin{examplebox}
	\noindent\textbf{\textcolor{teal}{Example \#A7:} \\  Factual Prompt (Science)}
	
	\vspace{2mm}
	\noindent\textbf{Prompt:}
	
	{\assamesefont QCD-ৰ সম্পূৰ্ণ নাম কি আৰু ই কোন মৌলিক বলৰ বিষয়ে অধ্যয়ন কৰে?}
	
	\vspace{2mm}
	\noindent\textbf{Model Response:}
	
	{	\assamesefont QCD-ৰ সম্পূৰ্ণ নাম হৈছে কোৱাণ্টাম ক্ৰ'ম’ডাইনেমিক্স (Quantum Chromodynamics) আৰু ই শক্তিশালী মৌলিক বলৰ অন্তৰ্গত কোৱাৰ্ক কণাৰ আচৰণ অধ্যয়ন কৰে ।
	}
\end{examplebox}

\begin{examplebox}
	\noindent\textbf{\textcolor{teal}{Example \#A8:} \\  Factual Prompt (Environment)}
	
	\vspace{2mm}
	\noindent\textbf{Prompt:}
	
	{\assamesefont বনাঞ্চল ধ্বংস হ'লে পৰিৱেশত কি কি প্ৰভাৱ পৰে?}
	
	\vspace{2mm}
	\noindent\textbf{Model Response:}
	
	{	\assamesefont বনাঞ্চল ধ্বংস কৰিলে জলবায়ু পৰিৱৰ্তন, মাটিৰ খহনীয়া আৰু জৈৱ-বৈচিত্ৰ্যৰ অৱনতি ঘটে, যিয়ে মানুহৰ জীৱন ধাৰণৰ বাবে ভাবুকি কঢ়িয়াই আনে।
	}
\end{examplebox}

\begin{examplebox}
	\noindent\textbf{\textcolor{teal}{Example \#A9:} \\  Reasoning}
	
	\vspace{2mm}
	\noindent\textbf{Prompt:}
	
	{\assamesefont কোনো ঐতিহাসিক ঘটনাৰ বিষয়ে দুটা নিৰ্ভৰযোগ্য উৎসত বেলেগ তথ্য থাকিলে শুদ্ধ তথ্য নিৰ্ধাৰণ কৰিবলৈ কি কি পদক্ষেপ ল'ব লাগে?}
	
	\vspace{2mm}
	\noindent\textbf{Model Response:}
	
	{	\assamesefont এনে পৰিস্থিতিত প্ৰথমে দুয়োটা উৎসৰ বিতং বিশ্লেষণ কৰিব লাগে, যেনে—আলোচনী, পুৰণি কিতাপ বা চৰকাৰী নথি। তাৰ পিছত তথ্যৰ সত্যতা পৰীক্ষা কৰিবলৈ বিভিন্ন অনলাইন আৰ্কাইভত বিচাৰ চলাব লাগে।
	}
\end{examplebox}

\begin{examplebox}
	\noindent\textbf{\textcolor{teal}{Example \#A10:} \\  Reasoning}
	
	\vspace{2mm}
	\noindent\textbf{Prompt:}
	
	{\assamesefont এটা AI মডেলৰ ওচৰত কোনো প্ৰশ্নৰ নিৰ্ভৰযোগ্য তথ্য নাথাকিলে ই কেনেদৰে উত্তৰ দিয়া উচিত?}
	
	\vspace{2mm}
	\noindent\textbf{Model Response:}
	
	{	\assamesefont এআই মডেল এটাই সঠিক উত্তৰ দিবলৈ তথ্যৰ ওপৰত নিৰ্ভৰ কৰে। যদি প্ৰশ্ন কৰা বিষয়ত পৰ্যাপ্ত তথ্য নাথাকে, তেন্তে ই 'unknown' বুলি ক'ব পাৰে।
	}
\end{examplebox}

\begin{examplebox}
	\noindent\textbf{\textcolor{teal}{Example \#A11:} \\  Cross-Lingual Adaptation for Romanized Assamese (Science)}
	
	\vspace{2mm}
	\noindent\textbf{Prompt:}
	
	{\assamesefont Nuclear fission aru nuclear fusionor majot mukhyo parthokyo ki?}
	
	\vspace{2mm}
	\noindent\textbf{Model Response:}
	
	{	\assamesefont পাৰমাণৱিক বিভাজন আৰু পাৰমাণৱিক সংযোজনৰ প্ৰধান পাৰ্থক্যটো হ'ল যে প্ৰথমটোত এটা গধুৰ নিউক্লিয়াছ বা পৰমাণু বিভাজন কৰা হয়, আনহাতে দ্বিতীয়টোত দুটা লঘু নিউক্লিয়াছ একত্ৰিত কৰা হয় যিটোৱে শক্তিৰ মুক্তি ঘটায়।
	}
\end{examplebox}

\begin{examplebox}
	\noindent\textbf{\textcolor{teal}{Example \#A12:} \\  Cross-Lingual Adaptation for Romanized Assamese (Assamese Literature)}
	
	\vspace{1mm}
	\noindent\textbf{Prompt:}
	
	{\assamesefont Lakshminath Bezbarua kun karone Axomiya xahityot bikhyato?}
	
	\vspace{1mm}
	\noindent\textbf{Model Response:}
	
	{	\assamesefont লক্ষ্মীনাথ বেজবৰুৱা অসমীয়া সাহিত্যত তেওঁৰ বহুমুখী প্ৰতিভাৰ বাবে বিখ্যাত আছিল। তেওঁ কবিতা, গল্প, নাটক, প্ৰবন্ধ আদি সকলো ক্ষেত্ৰতে অৱদান আগবঢ়াই অসমীয়া সাহিত্যক সমৃদ্ধ কৰিছিল।
	}
\end{examplebox}

\begin{examplebox}
	\noindent\textbf{\textcolor{teal}{Example \#A13:} \\  Cross-Lingual Adaptation for Romanized Assamese (Technology)}
	
	\vspace{1mm}
	\noindent\textbf{Prompt:}
	
	{\assamesefont Open-source software buli ki buja jai?}
	
	\vspace{1mm}
	\noindent\textbf{Model Response:}
	
	{	\assamesefont অপেন ছোর্স চফটৱেৰ হৈছে এনে এক প্ৰযুক্তি, য'ত ইয়াৰ উৎস ক'ড বা সূত্ৰসমূহ মুক্তভাৱে উপলব্ধ থাকে। ব্যৱহাৰকাৰীসকলে এই ক'ডৰ ওপৰত সংশোধন কৰিব পাৰে, ব্যৱহাৰ কৰিব পাৰে আৰু নিজৰ কামতো প্ৰয়োগ কৰিব পাৰে। ই বন্ধ ক'ডৰ বিপৰীতে এক সম্পূৰ্ণ বিপৰীত ধাৰণা।
	}
\end{examplebox}

\begin{examplebox}
	\noindent\textbf{\textcolor{teal}{Example \#A14:} \\  Reasoning}
	
	\vspace{1mm}
	\noindent\textbf{Prompt:}

	{\assamesefont সোমবাৰে গণিতৰ ক্লাছ হয়। মঙ্গলবাৰে বিজ্ঞানৰ ক্লাছ হয়। বিজ্ঞানৰ পিছৰ দিনা কলা (Art) ক্লাছ হয়। তেন্তে কলা ক্লাছ কি বাৰে হয়?}
	
	\vspace{1mm}
	\noindent\textbf{Model Response:}
	
	{	\assamesefont হয়, মঙলবাৰৰ পিছৰ দিনা অৰ্থাৎ বুধবাৰে কলাৰ ক্লাছ হয়।
	}
\end{examplebox}

\begin{examplebox}
	\noindent\textbf{\textcolor{teal}{Example \#A15:} \\  Reasoning}
	
	\vspace{1mm}
	\noindent\textbf{Prompt:}
	
	{\assamesefont এটা AI মডেলৰ ওচৰত কোনো প্ৰশ্নৰ নিৰ্ভৰযোগ্য তথ্য নাথাকিলে ই কেনেদৰে উত্তৰ দিয়া উচিত?}
	
	\vspace{1mm}
	\noindent\textbf{Model Response:}
	
	{	\assamesefont এআই মডেল এটাই সঠিক উত্তৰ দিবলৈ তথ্যৰ ওপৰত নিৰ্ভৰ কৰে। যদি প্ৰশ্ন কৰা বিষয়ত পৰ্যাপ্ত তথ্য নাথাকে, তেন্তে ই 'unknown' বুলি ক'ব পাৰে।
	}
\end{examplebox}

\begin{examplebox}
	\noindent\textbf{\textcolor{teal}{Example \#A16:} \\  Mathematical Reasoning}
	
	\vspace{1mm}
	\noindent\textbf{Prompt:}
	
	{\assamesefont 500 tokar 20\% kiman hoy?}
	
	\vspace{1mm}
	\noindent\textbf{Model Response:}
	
	{	\assamesefont ৫০০ টকাৰ ২০\% হৈছে ১০০ টকা।
	}
\end{examplebox}

\newpage
\subsection*{A.2 Long Answer Prompt-Response Examples / Creative Writting Examples}
The following qualitative examples illustrate the performance of AssameseLM-E2B-it on long-form creative writing tasks, highlighting its ability to maintain Assamese morphosyntactic stability over extended contexts without collapsing into sister languages.

\begin{examplebox}
	\noindent\textbf{\textcolor{teal}{Example \#B1:} \\  Assamese Creative Writing}
	
	\vspace{2mm}
	\noindent\textbf{Prompt:}
	
	{\assamesefont Write an essay on কৃত্ৰিম বুদ্ধিমত্তাই ভৱিষ্যতৰ শিক্ষা ব্যৱস্থাত কেনে প্ৰভাৱ পেলাব পাৰে?}
	
	\vspace{2mm}
	\noindent\textbf{Model Response:}
	
	{	\assamesefont
\# কৃত্ৰিম বুদ্ধিমত্তা আৰু ভৱিষ্যতৰ শিক্ষাৰ প্ৰত্যাহ্বান    \\ \\
\#\# ভূমিকা   

বৰ্তমানৰ দ্ৰুত পৰিৱৰ্তনশীল পৃথিৱীত মানৱ জাতিৰ ভৱিষ্যত নিৰ্ধাৰণত প্ৰযুক্তিৰ ভূমিকা অপৰিসীম। বিশেষকৈ, কৃত্ৰিম বুদ্ধিমত্তাৰ (AI) উত্থানে শৈক্ষিক ব্যৱস্থাৰ মূল ভেটি সলনি কৰিবলৈ গৈ আছে। ছাত্ৰ-ছাত্ৰীসকলে এই নতুন প্ৰযুক্তিক কেনেদৰে গ্ৰহণ কৰিব পাৰে বা ইয়াৰ পৰা উদ্ভৱ হোৱা প্ৰত্যাহ্বানৰ সন্মুখীন হ'ব পাৰে, সেয়া আজিৰ চিন্তাৰ বিষয় হৈ পৰিছে।     \\

\#\#\# AI-ৰ দ্বাৰা পাঠ্যক্ৰমৰ পৰিৱৰ্তন   \\

\#\#\# শিক্ষণৰ ধৰণ সলনি    

এআই-য়ে ছাত্ৰ-ছাত্ৰীক কেৱল তথ্য যোগান ধৰাই নহয়, বৰঞ্চ তেওঁলোকক প্ৰশ্ন কৰাৰ পদ্ধতি, সমস্যা সমাধানৰ পদ্ধতি আৰু সৃজনীমূলক চিন্তাধাৰাৰ ক্ষেত্ৰতো সহায় কৰিব। ব্যক্তিগতকৰণ কৰা পাঠ্যক্ৰম প্ৰস্তুত কৰাত এআইৰ ব্যৱহাৰ বৃদ্ধি পাব, যাৰ ফলত ছাত্ৰ-ছাত্ৰীয়ে নিজৰ গতি অনুসৰি শিকিবলৈ সুযোগ পাব। কিন্তু ডিজিটেল ডিভাইচসমূহৰ অত্যধিক ব্যৱহাৰৰ ফলত শ্ৰেণীকোঠাৰ বাহিৰতো একাকীত্ব আৰু মানসিক চাপৰ সৃষ্টি হোৱাৰ সম্ভাৱনা থাকে।     \\

\#\#\# কৰ্মসংস্থান সৃষ্টিৰ ক্ষেত্ৰত প্ৰত্যাহ্বান    \\

\#\#\# এআই-এ মানুহৰ চাকৰিৰ গুৰুত্ব হ্ৰাস কৰিব বুলি আশংকা কৰা হয়। বৰ্তমান সময়ত বহু কাম যেনে অনুবাদ কৰা, ডাটা এণ্ট্ৰি কৰা আদি কামবোৰ স্বয়ংক্ৰিয়ভাৱে সম্পাদন হ'ব। যদিও ই কিছুমান পুৰণি বৃত্তিকে বিলুপ্ত কৰিব, নতুন ধৰণৰ জ্ঞান, সৃষ্টিশীলতা আৰু নেতৃত্বৰ প্ৰয়োজনীয়তাহে অধিক গুৰুত্বপূৰ্ণ। শিক্ষাত যান্ত্ৰিক কামৰ পৰিৱৰ্তে মানৱীয় দক্ষতাৰ ওপৰত জোৰ দিব লাগিব।     \\

\#\# সামাজিক আৰু নৈতিক দিশসমূহ    

\#\#\# তথ্যৰ অপব্যৱহাৰ আৰু ভুৱা খবৰ    \\

এআই-এ ইণ্টাৰনেটৰ জৰিয়তে বিশাল পৰিমাণৰ তথ্যৰ আদান-প্ৰদানত সহায় কৰে, যিটো ইতিবাচক কথা। অৱশ্যে, ইয়াত ভুল তথ্য বা ভুৱা খবৰৰ সংক্ৰমণ ৰোধ কৰাটো অতি প্ৰয়োজনীয়। ছাত্ৰ-ছাত্ৰীসকলৰ মাজত তথ্যৰ বিচাৰ ক্ষমতা শক্তিশালী কৰিলে সমাজত শান্তি বজাই ৰখাত সহায়ক হ'ব।    \\

\#\#\# সমতা আৰু অন্তৰ্ভুক্তিৰ প্ৰসাৰ    \\

\#\#\# প্ৰাথমিক পৰ্যায়ত AI প্ৰয়োগৰ সঠিক পৰিকল্পনা অত্যন্ত জৰুৰী। দুখীয়া অঞ্চলৰ ছাত্ৰ-ছাত্ৰী আৰু গ্ৰাম্য অঞ্চলত এতিয়াও ইন্টাৰনেটৰ সীমিত প্ৰৱেশাধিকাৰ দেখা যায়, যি তেওঁলোকৰ বাবে শিক্ষাৰ অধিকাৰ প্ৰাপ্তি কঠিন কৰি তোলে। চৰকাৰে বিনামূলীয়া শিক্ষা প্ৰদান ব্যৱস্থা বিকশিত কৰি এআইৰ সুফল যাতে সকলোৰে ওচৰলৈ যায় তাৰ প্ৰতি লক্ষ্য ৰাখিব লাগে।      \\

\par সামৰণিত ক'ব পাৰি যে কৃত্ৰিম বুদ্ধিমত্তাৰ আশীৰ্বাদ হিচাপে শিক্ষাই এক নতুন দিগন্ত দেখিছে; আনহাতে ইয়াক নিয়ন্ত্ৰণ অবিহনে প্ৰয়োগ কৰিলে আমাৰ সমাজ গাঁথনি ভাঙি যোৱাৰ ভয় আছে। গতিকে আমি সকলোৱে মিলি এখন সুৰক্ষিত আৰু প্ৰগতিশীল ভৱিষ্যতৰ পথ বাছি ল'ব লাগিব।
	}
\end{examplebox}

\begin{examplebox}
	\noindent\textbf{\textcolor{teal}{Example \#B2:} \\  Assamese Creative Writing}
	
	\vspace{2mm}
	\noindent\textbf{Prompt:}
	
	{\assamesefont বানপীড়িত পৰিয়ালসমূহৰ বাবে সাহায্য প্ৰদানৰ অনুৰোধ জনাই জিলা আয়ুক্তলৈ এখন আবেদন লিখা। }
	
	\vspace{2mm}
	\noindent\textbf{Model Response:}
	
	{	\assamesefont 
	প্ৰতি,\\
	জিলা আয়ুক্ত,\\
	কামৰূপ মহানগৰ জিলা প্ৰশাসন, গুৱাহাটী।  \\
	
	তাৰিখ: ২৪ মে', ২০২৪ 
	
	বিষয়: বান বিধ্বস্ত পৰিয়ালৰ বাবে সাহাৰ্য্য প্ৰদান কৰাৰ আবেদন।   \\
	
	মহোদয়, 
	
	সন্মানপূৰ্বক বিনীত নিবেদন এই যে, যোৱা দুদিন ধৰি হোৱা নেৰানেপেৰা বৰষুণৰ ফলত আমাৰ অঞ্চলটোত এক ভয়াৱহ পৰিস্থিতিৰ সৃষ্টি হৈছে। নদীৰ পানী বাঢ়ি অহাৰ ফলত বহু লোকৰ ঘৰ-দুৱাৰ ধ্বংস হৈছে আৰু হাজাৰ হাজাৰ লোক গৃহহীন হৈ পৰিছে। এই দুৰ্যোগৰ সময়ত আটাইতকৈ বেছি ক্ষতিগ্ৰস্ত হৈছে দৰিদ্ৰ লোকসকল আৰু শিশুসকল।   \\
	
	গতিকে, আপোনাৰ ওচৰত আমাৰ বিনীত প্ৰাৰ্থনা, অনুগ্ৰহ কৰি বানাক্ৰান্ত অঞ্চলৰ লোকসকলৰ বাবে প্ৰয়োজনীয় খাদ্য, বিশুদ্ধ খোৱাপানী আৰু কাপোৰ-কানি আদি সামগ্ৰী বিতৰণ কৰিবলৈ প্ৰশাসনক নিৰ্দেশ প্ৰদান কৰিব। আমি নিশ্চিত যে চৰকাৰৰ সহায়ৰ হাত পাই সকলো লোক উপকৃত হ'ব।  \\ 
	
	ধন্যবাদেৰে, \\
	স্থানীয় ৰাইজৰ হৈ,  \\
	ৰঞ্জিত দাস।
	}
\end{examplebox}

\begin{examplebox}
	\noindent\textbf{\textcolor{teal}{Example \#B3:} \\  Assamese Creative Writing}
	
	\vspace{2mm}
	\noindent\textbf{Prompt:}
	
	{\assamesefont পূৰ্বৰ কৰ্মস্থলীৰ পৰা অভিজ্ঞতাৰ প্ৰমাণ-পত্ৰ বিচাৰি এখন আবেদন লিখা। }
	
	\vspace{2mm}
	\noindent\textbf{Model Response:}
	
	{	\assamesefont 
প্ৰতি, \\
শাখা প্ৰবন্ধক, \\
এ বি চি প্ৰাইভেট লিমিটেড, গুৱাহাটী।  \\

তাৰিখ: ২৪ অক্টোবৰ, ২০২৪ 

বিষয়: অভিজ্ঞতামূলক প্ৰমাণ-পত্ৰ (Experience Certificate) প্ৰদানৰ বাবে আবেদন।  \\

মহোদয়, \\
মই আপোনাৰ প্ৰতিষ্ঠানৰ এগৰাকী কৰ্মচাৰী হিচাপে বিগত পাঁচ বছৰ ধৰি সেৱা আগবঢ়াই আহিছোঁ। মোৰ মূল দায়িত্ব আছিল—কমাণ্ড এণ্ড কণ্ট্ৰল চাৰ্ভে (CPC) আৰু ডাটা এনালাইছিছৰ কামত কম্পিউটাৰ ছফ্টৱেৰ ব্যৱহাৰ কৰা। এই সময়ছোৱাত মই বহু জটিল প্ৰকল্পৰ সফল ৰূপায়ণত অংশ লৈছোঁ।   \\

মোৰ কাৰ্যকাল শেষ হোৱাৰ লগে লগে মোক চাকৰি জীৱনৰ অভিজ্ঞতা প্ৰতিফলিত কৰিব পৰাকৈ এটা অভিজ্ঞতামূলক প্ৰমাণ-পত্ৰ প্ৰদান কৰিবলৈ অনুৰোধ জনালোঁ। আশা কৰোঁ আপুনি মোৰ সেৱাৰ মূল্যায়ন কৰি প্ৰয়োজনীয় ব্যৱস্থা গ্ৰহণ কৰিব।   \\

ধন্যবাদেৰে, \\
আপোনাৰ বিশ্বস্ত,  \\
ৰাজেশ বৰা।
	}
\end{examplebox}

\subsection*{A.3 RAG Output}
The following qualitative examples illustrate the performance of AssameseLM-E2B-it on RAG tasks, highlighting critical differences in context adherence, script fidelity, the mitigation of hallucinations and RAG refusal.

\begin{examplebox}
	\noindent\textbf{\textcolor{teal}{Example \#C1:} \\  Native Assamese RAG}
	
	\vspace{2mm}
	\noindent\textbf{Prompt:}
	
	{\assamesefont 
	Context:
	[1] কাজিৰঙা ৰাষ্ট্ৰীয় উদ্যান এশিঙীয়া গঁড়ৰ বাবে বিখ্যাত।
	[2] মাজুলী হৈছে ব্ৰহ্মপুত্ৰ নদীৰ মাজত অৱস্থিত এখন বৃহৎ নদীদ্বীপ।
	[3] মানাহ ৰাষ্ট্ৰীয় উদ্যান বাক্সা আৰু চিৰাং জিলাত অৱস্থিত।
	\\ \\
	Question: মাজুলী কোনখন নদীৰ মাজত অৱস্থিত?
	}
	
	\vspace{2mm}
	\noindent\textbf{Model Response:}
	
	{	\assamesefont 
		মাজুলী হৈছে ব্ৰহ্মপুত্ৰ নদীৰ মাজত অৱস্থিত এখন বৃহৎ নদীদ্বীপ [2].
	}
\end{examplebox}

\begin{examplebox}
	\noindent\textbf{\textcolor{teal}{Example \#C2:} \\  Cross-lingual RAG}
	
	\vspace{2mm}
	\noindent\textbf{Prompt:}
	
	{\assamesefont 
	Context:
	[1] Nameri National Park is located in the Sonitpur district and is famous for the White-winged Wood Duck.
	[2] The Golden Langur Conservation Breeding Programme is successfully running at the Assam State Zoo in Guwahati.
	[3] Pobitora Wildlife Sanctuary has the highest density of one-horned rhinos in the world.
	\\ \\
	Question: State Zoo kot ase aaru tat kihor breeding programme soli ase?
	}
	
	\vspace{2mm}
	\noindent\textbf{Model Response:}
	
	{	\assamesefont অসম ৰাজ্যিক চিৰিয়াখানাত সোণালী বান্দৰৰ সংৰক্ষণ প্ৰজনন কাৰ্যসূচী চলি আছে [2]।
	}
\end{examplebox}

\begin{examplebox}
	\noindent\textbf{\textcolor{teal}{Example \#C3:} \\  Cross-lingual RAG}
	
	\vspace{2mm}
	\noindent\textbf{Prompt:}
	
	{\assamesefont 
		Context:
		[1] ড° ভূপেন হাজৰিকা অসমৰ এগৰাকী সুধাকণ্ঠ নামেৰে পৰিচিত কিংবদন্তি গায়ক।
		[2] He was awarded the Bharat Ratna posthumously in the year 2019.
		\\ \\
		Question: Who is known as Sudhakantha and in which year did he receive the Bharat Ratna?
	}
	
	\vspace{2mm}
	\noindent\textbf{Model Response:}
	
	{	\assamesefont 
		Dr. Bhupen Hazarika is known as Sudhakantha [1], and he received the Bharat Ratna in the year 2019 [2].
	}
\end{examplebox}

\begin{examplebox}
	\noindent\textbf{\textcolor{teal}{Example \#C4:} \\  Cross-lingual RAG}
	
	\vspace{2mm}
	\noindent\textbf{Prompt:}
	
	{\assamesefont 
		Context:
		[1] The Ahom kingdom was established by Chaolung Sukaphaa in the year 1228.
		[2] The historic Battle of Saraighat was fought in 1671 between the Mughal Empire and the Ahom Kingdom.
		\\ \\
		Question: আহোম সাম্ৰাজ্য কেতিয়া প্ৰতিষ্ঠা হৈছিল আৰু শৰাইঘাটৰ ৰণ কোন চনত হৈছিল?
	}
	
	\vspace{2mm}
	\noindent\textbf{Model Response:}
	
	{	\assamesefont 
		আহোম সাম্ৰাজ্য ১২২৮ চনত প্ৰতিষ্ঠা কৰা হৈছিল [1], আৰু শৰাইঘাটৰ ৰণ ১৬৭১ চনত হৈছিল [2]।
	}
\end{examplebox}

\begin{examplebox}
	\noindent\textbf{\textcolor{teal}{Example \#C5:} \\  RAG Refusal}
	
	\vspace{2mm}
	\noindent\textbf{Prompt:}
	
	{\assamesefont 
		Context:
		[1] ৰঙালী বিহু বা ব'হাগ বিহু অসমৰ মূল উৎসৱ আৰু ই নৱবৰ্ষৰ আৰম্ভণিকা সূচায়।
		[2] অম্বুবাচী মেলা প্ৰতি বছৰে আহাৰ মাহত গুৱাহাটীৰ কামাখ্যা মন্দিৰত অনুষ্ঠিত হয়।
		\\ \\
		Question: জোনবিল মেলা কোনখন জিলাত অনুষ্ঠিত হয়?
	}
	
	\vspace{2mm}
	\noindent\textbf{Model Response:}
	
	{	\assamesefont 
		ক্ষমা কৰিব, প্ৰদান কৰা তথ্যত ইয়াৰ উত্তৰ উপলব্ধ নহয়।
	}
\end{examplebox}

\subsection*{A.4 Examples where The AssameseLM-E2B-it  Model Generates Bad Output}
AssameseLM-E2B-it  is prone to hallucinations and can present distorted facts as D1 \& D2.

\begin{examplebox}
	\noindent\textbf{\textcolor{teal}{Example \#D1:} \\  Factual Prompt-Response hallucination}
	
	\vspace{2mm}
	\noindent\textbf{Prompt:}
	
	{\assamesefont প্ৰ'টন আৰু নিউট্ৰন কোনবোৰ মৌলিক কণাৰ দ্বাৰা গঠিত?  }
	
	\vspace{2mm}
	\noindent\textbf{Model Response:}
	
	{	\assamesefont প্ৰ'টন এটা উৎপন্ন হোৱা মৌলিক কণা নহয়, ই নিজেই এটা প্ৰ'টন। আনহাতে, নিউট্ৰন হৈছে এটা মৌলিক কণা যি ইলেক্ট্ৰনৰ সৈতে একে ভৰৰ কিন্তু বিপৰীত আধানযুক্ত।
	}
\end{examplebox}

\begin{examplebox}
	\noindent\textbf{\textcolor{teal}{Example \#D2:} \\  Factual Prompt-Response hallucination}
	
	\vspace{2mm}
	\noindent\textbf{Prompt:}
	
	{\assamesefont শব্দই শূন্যস্থানৰ মাজেৰে কিয় গতি কৰিব নোৱাৰে?}
	
	\vspace{2mm}
	\noindent\textbf{Model Response:}
	
	{	\assamesefont শব্দই শূন্যস্থানৰ মাজেৰে গতি কৰিব নোৱাৰে কাৰণ ই এক ভৌতিক বস্তু নহয়, বৰঞ্চ এটা সংকেতহে।
	}
\end{examplebox}

 \section{Acknowledgements}

The authors acknowledge the use of Google's Gemini and OpenAI's ChatGPT as AI-assisted tools during the preparation of this manuscript. These tools were utilized to improve language, refine the presentation of ideas, and enhance the overall readability and coherence of the manuscript. The authors take full responsibility for the scientific content, interpretations, and conclusions presented in this work.

\bibliographystyle{plainnat}
\bibliography{assamese_lm_references}

\end{document}